\documentclass[10pt,twocolumn,letterpaper]{article}

\usepackage{iccv}
\usepackage{times}
\usepackage{epsfig}
\usepackage{graphicx}
\usepackage{amsmath}
\usepackage{amssymb}
\usepackage{url}

\usepackage[pagebackref=true,breaklinks=true,letterpaper=true,colorlinks,bookmarks=false]{hyperref}

\iccvfinalcopy 

\def\iccvPaperID{133} 

\ificcvfinal\fi

\usepackage{booktabs}
\usepackage{xspace}
\renewcommand{\vec}[1]{\mathbf{\boldsymbol{#1}}}
\newcommand{\mat}[1]{\mathbf{#1}}

\newcommand\equationname{Eq.}
\newcommand\sectionname{Sect.}

\ifdefined\eg
\else
  \newcommand\eg{\textit{e.g.},\xspace}
\fi

\ifdefined\ie
\else
  \newcommand\ie{\textit{i.e.},\xspace}
\fi

\ifdefined\cf
\else
  \newcommand\cf{\textit{cf.}\xspace}
\fi

\newcommand\landau{\mathcal{O}}

\newcommand\inputsingle{\vec{x}}

\newcommand{\infosingle}{\vec{v}}

\newcommand\labelsingle{y}

\newcommand\dataset{\mathcal{D}}

\newcommand\dimension{D}
\newcommand\noe{n}

\usepackage{upgreek}

\newcommand\lagrangeDual{g}

\newcommand\identityMatrix[1]{\mat{I}}

\newcommand\optimizationProblem[5]{
	\begin{equation}
	\label{#1}
	\begin{aligned}
	& \underset{#3}{#2}
	& & #4 \\
	& \text{s.t.}
	& & #5 \enspace.
	\end{aligned}
	\end{equation}
}

\newcommand\dainputsingle{\tilde{\vec{x}}}

\newcommand\dalabelsingle{\tilde{y}}
\newcommand\dadataset{\tilde{\mathcal{D}}}

\newcommand\datransform{\mat{W}}

\newcommand\myparagraph[1]{\vspace{-10pt}\paragraph{#1}}

\begin{document}

\title{Towards Adapting ImageNet to Reality: Scalable Domain Adaptation with Implicit Low-rank Transformations}

\author{
Erik Rodner$^{1,2,}$\thanks{both authors contributed equally}
\quad Judy Hoffman$^{1,*}$ \quad Jeff Donahue$^{1}$\\
Trevor Darrell$^{1}$ \quad Kate Saenko$^{3}$\\
$^1$ICSI \& EECS UC Berkeley, $^2$University of Jena, $^3$UMass Lowell
}

\newcommand\todo[1]{\textcolor{red}{#1}}
\maketitle

\begin{abstract}
Images seen during test time are often not from the same distribution as images used for learning.
This problem, known as domain shift, occurs when training classifiers from object-centric internet image databases and trying to apply them directly to scene understanding tasks.
The consequence is often severe performance degradation and is one of the major barriers for the application of classifiers in real-world systems.
In this paper, we show how to learn transform-based domain adaptation classifiers in a scalable manner.
The key idea is to exploit an implicit rank constraint, originated from a max-margin domain adaptation formulation, to make optimization tractable.
Experiments show that the transformation between domains can be very efficiently learned from data and easily applied to new categories.
This begins to bridge the gap between large-scale internet image collections and object images captured in everyday life environments.
\end{abstract}

\section{Introduction}

\begin{figure}[t]
    \includegraphics[width=\linewidth]{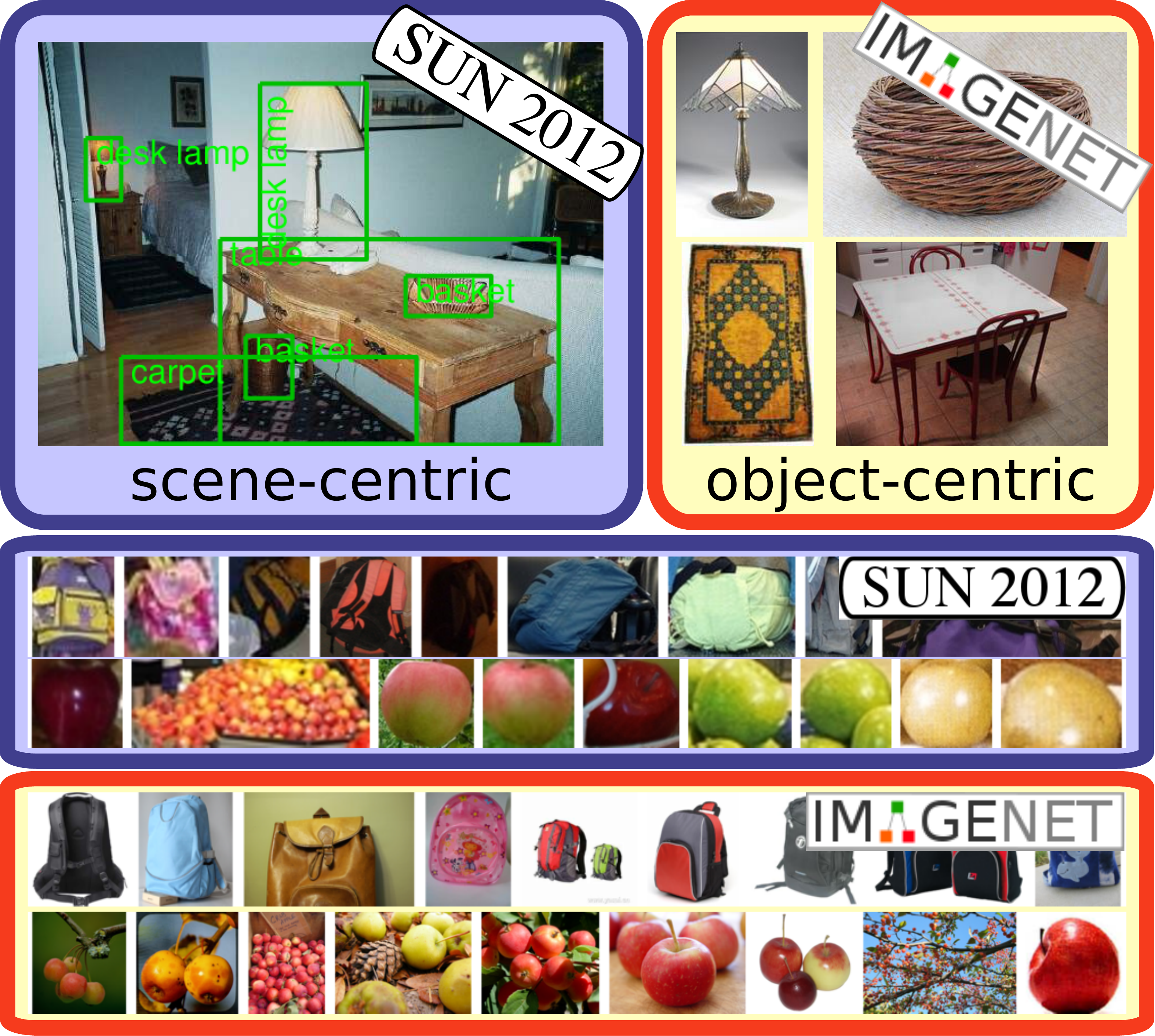}
    \caption{Dataset bias of ImageNet and the SUN2012 database shown for an indoor scene and for the categories \emph{backpack} and \emph{apple} on a bounding box level.}
    \label{fig:bias}
  \end{figure}

 Learning from huge datasets comprised of millions of images is one of the most promising directions towards closing the gap between human and machine
 visual recognition abilities. There has been tremendous success in the area of large-scale visual recognition~\cite{Deng09:ILS} allowing for learning of tens of thousands of visual categories. However, in parallel, researchers have discovered the bias induced by current image databases and that performing visual recognition tasks across domains cripples performance~\cite{ref:Efros-dataset-bias-cvpr2011}. Although this is especially common for smaller datasets, like Caltech-101 or the PASCAL VOC datasets~\cite{ref:Efros-dataset-bias-cvpr2011}, the way large image databases are collected (typically using internet search engines) also introduces an inherent bias. This can be seen for example when comparing object images of the ImageNet~\cite{Deng09:ILS} and SUN2012 database~\cite{Xiao10:sun} in \figurename~\ref{fig:bias}, where the ``object-centric'' data of ImageNet is of high resolution with centered objects as well as sometimes artificial backgrounds, and the SUN2012 objects are part of scene images leading to blurred appearances with a large degree of occlusion and truncation.

 Transform-based domain adaptation overcomes the bias by learning a transformation between datasets. In contrast to classifier adaptation~\cite{Aytar11:TRM,ref:yang,BergamoTorresani10,ref:khosla_eccv12}, learning a transformation between feature spaces directly allows us to perform adaptation even for (new) categories that are not present in both datasets. Especially for large-scale recognition with a large number of categories, this is a crucial benefit, because we can learn category models for all the categories in a given source domain also in the target domain. Transformations can be learned in an unsupervised manner~\cite{Kulis11:WYS} or by using the labels present in both domains to maximize the margin of the classifier on the source and transformed target data~\cite{Hoffman13:ELD,ref:duan_icml12}.

 In this paper, we introduce a novel optimization method that enables transform-learning and associated domain adaptation methods to scale to ``big data''. 
 We do this by a novel re-formulation of the optimization in~\cite{Hoffman13:ELD} as direct dual coordinate descent and by exploiting an implicit rank constraint.
 Although we learn a linear transformation between domains, which has a quadratic size in the number of features used, our algorithm needs only a linear number of operations in each iteration in both feature dimensions (source and target domain) as well as the number of training examples. This is an important benefit compared to other methods that need to run in kernel space~\cite{Kulis11:WYS,ref:duan_icml12} to overcome the high dimensionality of the transformation, a strategy impossible to apply for large-scale settings.
  The obtained scalability of our method is crucial as it allows the use of transform-based domain adaptation for datasets with a large number of categories and examples, settings in which previous techniques~\cite{Kulis11:WYS,ref:duan_icml12,Hoffman13:ELD} were unable to run in reasonable time. Our experiments on different datasets show the various advantages of transform-based methods, such as generalization to new categories or even handling domains with different feature types.

\section{Related Work}
\vspace{-.2cm}
  For the task of domain adaptation, two different sets of data are typically considered, the source and the target domain, which are drawn from similar but distinct distributions $p(\inputsingle)$ and $p(\dainputsingle)$. The goal is to transfer knowledge from the source domain to the target domain. In the following, we briefly review related work done in the areas of domain adaptation as well as transfer learning. 
  Although transfer learning~\cite{Pan10STL} considers a change of the conditional distribution $p(\labelsingle \,|\, \inputsingle)$  rather than a change of the data distribution $p(\inputsingle)$ as in domain adaptation, the methods in both areas often use similar principles and ideas.

    Domain adaptation can be applied at different levels of the machine learning pipeline. For example, the adaptive SVM method~\cite{ref:yang}
    combines a target classifier and an existing source classifier by linear combination of their continuous outputs. This is related to adding a new regularization term to
    the SVM objective that forces the target SVM hyperplane parameter to be close to the source hyperplane~\cite{Yang07:ASC}.
    Aytar and Zisserman~\cite{Aytar11:TRM} showed the importance of using a scale-invariant similarity measure for this regularization term.
    Furthermore, the authors of \cite{BergamoTorresani10} proposed a combination of target, source and transductive SVM. 
    More recently, Khosla~\etal~\cite{ref:khosla_eccv12} introduced a method to jointly learn a ``visual world model'' common across all domains in combination with an additive bias term for each individual domain.

    In general, classifier adaptation methods are often limited to cases where labeled training data is given for every class in the source as well as in the target domain. However, we often have a source domain with not only more training examples but also more labeled categories available. Exploiting all the information and learning visual classifiers for new categories in the target domain is possible with metric or transformation-based methods. 

  Another line of work was started by Gopalan~\etal~\cite{Gopalan11:DAO}, who introduced domains as points on a manifold of subspaces. To perform domain adaptation, features are mapped to the subspaces induced by the geodesic from the source to the target domain. This yields several intermediate representations of the input data that can be used for learning a classifier. Gong~\etal~\cite{ref:gong12_gfk} showed how to circumvent sampling only a finite number of subspaces by expressing the representation as a kernel. 
      In contrast, Tommasi~\etal~\cite{Tommasi12:BDB} tackled the domain adaptation problem by learning a shared subspace capturing domain-invariant properties of the categories. Learning for a new dataset is then done by learning an additional domain-specific transformation of the data.

    The work of Saenko~\etal~\cite{Saenko10:AVC} was one of the earliest papers to investigate domain adaptation challenges in visual recognition. The key idea of their work is to apply metric learning techniques that allow for estimating a category-independent metric which related target and source examples, and can be used in a nearest neighbor classifier.
    Kulis~\etal~\cite{Kulis11:WYS} extended their work to asymmetric transformations and metrics using a Frobenius norm regularizer.
    A major bottleneck of their approach is the number of instance (linear) constraints, one for each pair of source and target examples, that need to be considered during optimization and the fact that transforms are learned independently of loss. Therefore, Hoffman~\etal~\cite{Hoffman13:ELD} recently showed how to jointly learn a transformation together with SVM parameters in a max-margin framework, which reduces the number of constraints to the number of categories.
    The linear transformation was quadratic in the feature dimensionality, and the kernelization as used by \cite{Hoffman13:ELD,Kulis11:WYS} was quadratic in the number of training examples. 
    This scales poorly with very large data, and as we show in the experiments section is intractable for even modestly large-scale data.

\section{Scalable Transformation Learning}
We introduce a method for learning a transformation, which is easy to apply, implement, and can be combined with other large-scale architectures.
Our new scalable method can be applied to supervised domain adaptation, where we are given source training examples $\dataset = \left\{ (\inputsingle_i, \labelsingle_i) \right\}_{i=1}^\noe$ and target examples $\dadataset = \left\{ (\dainputsingle_j, \dalabelsingle_j) \right\}_{j=1}^{\tilde{\noe}}$. 

Our goal is to learn a linear transformation $\datransform \dainputsingle$ mapping a target training data point $\dainputsingle$ to the source domain. The transformation is learned through an optimization framework which introduces linear constraints between transformed target training points and information from the source and thus generalizes the methods of \cite{Saenko10:AVC,Kulis11:WYS,Hoffman13:ELD}. 
To demonstrate the generality of our approach, we denote linear constraints in the source domain using hyperplanes $\infosingle_i \in \mathbb{R}^\dimension$ for $1 \leq i \leq m$. Let us denote with $\dalabelsingle_{ij}$ a scalar which represents some measure of intended similarity between $\infosingle_i$ and the target training data point $\dainputsingle_j$. With this general notation, we can express the standard transformation learning problem with slack variables as follows:
\optimizationProblem{eq:datransform}{\min}
                    {\datransform, \{ \vec{\eta} \}}
                    {\frac{1}{2} \| \datransform \|^2_F + \tilde{C} \sum\limits_{i=1,j=1}^{m,\tilde{\noe}} {(\eta_{ij} )}^p}
                    { \dalabelsingle_{ij} \left( \infosingle_i^T  \datransform \dainputsingle_j \right) \geq 1 - \eta_{ij}, \; \eta_{ij} \geq 0 \quad \forall i,j}
Note that this directly corresponds to the transformation learning problem proposed in \cite{Hoffman13:ELD}.
Previous transformation learning techniques~\cite{Saenko10:AVC, Kulis11:WYS, Hoffman13:ELD} used a Bregman divergence optimization technique~\cite{Kulis11:WYS}, which scales quadratically in the number of target training examples (kernelized version) or the number of feature dimensions (linear version). For the large-scale scenario considered in this paper, this is impractical due to the large number of target training examples and categories given, as well as the high dimensionality of the features. Therefore, we show in a new analysis both how to use dual coordinate descent for the optimization of $\datransform$ and that $\datransform$ has a low-rank structure, which can be exploited to allow for efficient optimization as verified in our experimental evaluation.

\subsection{Learning $\datransform$ with dual coordinate descent}
\newcommand\vectorize[1]{\text{vec}\left(#1\right)}

We now re-formulate \equationname~\eqref{eq:datransform} as a vectorized optimization problem suitable for dual coordinate descent that allows us to use efficient optimization techniques. 
We use $\vec{w} = \vectorize{\mat{\datransform}}$ to denote the vectorized version of a matrix $\mat{\datransform}$ obtained by concatenating the rows of the matrix into a single column vector. With this definition, we can write:
\begin{align}
\|\datransform\|_F^2 &= \|\vectorize{\datransform}\|_2^2 = \| \vec{w} \|^2_2 \\
\infosingle_i^T \datransform \dainputsingle_j &= \vec{w}^T \vectorize{ \infosingle_i  \cdot \dainputsingle_j^T } \enspace.
\end{align}
Let $\ell = m (j-1) + i$ be the index ranging over the target examples as well as the $m$ hyperplanes in the source domain, which we also denote as $\ell = (i,j)$ for convenience. We now define a new set of ``augmented'' features as follows:
\begin{align}
  \vec{d}_\ell &=  \vectorize{ \infosingle_i \cdot \dainputsingle_j^T } \in \mathbb{R}^{\dimension \times \tilde{\dimension}} \enspace,\\
  t_\ell &= \dalabelsingle_{ij} \enspace.
\end{align}
With these definitions, \equationname~\eqref{eq:datransform} is equivalent to a soft-margin SVM problem with training set $\left(  \vec{d}_\ell, t_\ell \right)_{\ell=1}^{\tilde{n} \cdot K}$. We exploit this result of our analysis by using and modifying the efficient coordinate descent solver proposed in \cite{Hsieh08:DCD}, which solves the SVM optimization problem in its dual form with respect to the dual variables $\alpha_\ell$:
\begin{align}
\label{eq:dual}
\min_{\vec{\alpha} \geq 0} \lagrangeDual(\vec{\alpha}) &= \frac{1}{2} \vec{\alpha}^T \bar{\mat{Q}} \vec{\alpha} - \vec{e}^T \vec{\alpha} \enspace.
\end{align}
We have considered the $L_2$-SVM formulation ($p=2$ in \equationname~\eqref{eq:datransform}), although our techniques presented in this paper also hold for the standard $L_1$-SVM case. The matrix $\mat{Q}$ is a regularized kernel matrix incorporating the labels, \ie $\bar{Q}_{\ell,\ell'} = t_\ell t_{\ell'} \, \vec{d}_\ell^T \vec{d}_{\ell'} + \lambda \,\delta\left[i=j\right]$ with $\lambda = \frac{1}{2\tilde{C}}$.
The key idea is to maintain and update $\vec{w}$ explicitly:
\begin{align}
\label{eq:representer}
  \vec{w} &= \sum\limits_{\ell=1}^{m \cdot \tilde{\noe}} \alpha_\ell \; t_\ell \; \vec{d}_\ell \enspace.
\end{align}
This dramatically reduces the computational complexity of the gradient computation in $\alpha_\ell$ compared to classical dual solvers commonly used for kernel SVM: 
\begin{align}
\label{eq:gradient}
\nabla_\ell\, \lagrangeDual(\vec{\alpha}) &= \labelsingle_i \cdot \vec{w}^T \vec{d}_\ell + \lambda\,\alpha_\ell - 1 \enspace,
\end{align}
which requires a number of operations linear in the dimensionality of the given (augmented) feature vectors $\vec{d}_\ell$. A single coordinate descent step can then be done by:
\begin{align}
\label{eq:alphaupdate}
\alpha_\ell &\leftarrow \max\left( 0, \alpha_\ell - \frac{\nabla_\ell\, \lagrangeDual (\vec{\alpha})}{\|\vec{d}_\ell\| + \lambda} \right)
\end{align}
in the same asymptotic time. Note that explicitly maintaining $\vec{w}$ is essential for easily computable coordinate descent steps; therefore, given the change $\triangle \alpha_\ell$ of the step, we have to update $\vec{w}$ so that \equationname~\eqref{eq:representer} is again fulfilled:
\begin{align}
\label{eq:wupdate}
\vec{w} &\leftarrow \vec{w} + \triangle \alpha_\ell \, t_\ell \, \vec{d}_\ell \enspace.
\end{align}
Whereas, for standard learning problems an iteration with only a linaer number of operations in the feature dimensionality already provides a sufficient speed-up, this is not the case when learning domain transformations $\datransform$. When the dimension of the source and target feature space is $\dimension$ and $\tilde{\dimension}$, respectively, the features $\vec{d}_\ell$ of the augmented training set have a dimensionality of $\dimension \cdot \tilde{\dimension}$, which is impractical for vision tasks with high-dimensional input features. For this reason, we show in the following how we can efficiently exploit an implicit low-rank structure of $\datransform$ for a small number of hyperplanes inducing the constraints.

\subsection{Implicit low-rank structure of the transform}

To derive a low-rank structure of the transformation matrix, 
let us recall \equationname~\eqref{eq:representer} in matrix notation:
\begin{align}
\datransform &= \sum\limits_{i=1,j=1}^{m,\tilde{\noe}} \alpha_{\ell}\; \infosingle_i \cdot \dainputsingle_j^T = \sum\limits_{i=1}^m \infosingle_i \left( \sum\limits_{j=1}^{\tilde{\noe}} \alpha_\ell \, \dainputsingle_j^T \right) \enspace.
\end{align}
Thus, $\datransform$ is a sum of $m$ dyadic products and therefore a matrix of at most rank $m$, with $m$ being the number of hyperplanes in the source used to generate constraints. 
Note that for our experiments, we use the MMDT method~\cite{Hoffman13:ELD}, for which the number of hyperplanes equals the number of object categories we seek to classify.
We can exploit the low rank structure by representing $\datransform$ indirectly using:
\begin{align}
\vec{\beta}_i &= \sum\limits_{j=1}^{\tilde{\noe}} \alpha_\ell \, \dainputsingle_j^T \enspace. 
\end{align}
This is especially useful when the number of categories is small compared to the dimension of the source domain, because $\left[ \vec{\beta}_1, \ldots, \vec{\beta}_m \right]$ only has a size of $m \times \tilde{\dimension}$ instead of $\dimension \times \tilde{\dimension}$ for $\datransform$. It also allows for very efficient updates with a computation time even independent of the number of categories.

First, with the given low-rank representation and the $\vec{\beta}_i$, we can easily speed up the scalar product in \equationname~\eqref{eq:gradient}:
\begin{align*}
  \vec{w}^T \vec{d}_\ell &= \infosingle_i^T \datransform \dainputsingle_j = \sum\limits_{i'=1}^m \infosingle_i^T \infosingle_{i'} \, \vec{\beta}_{i'}^T \dainputsingle_j = \sum\limits_{i'=1}^m \rho_{i,i'} \vec{\beta}_{i'}^T \dainputsingle_j \;,
\end{align*}
where the matrix $\mat{R} = (\rho_{i,i'}) \in \mathbb{R}^{m \times m}$ can be calculated in advance. Furthermore, we can cache $\vec{\beta}_{i'}^T \dainputsingle_j$, leading to a only a cost of $\mathcal{O}(\tilde{\dimension})$ (Details in \sectionname~\ref{sec:algodetails}).

The matrix $\mat{R}$ contains the correlations between hyperplanes and also shows the multi-task fashion of the approach: the $\vec{\beta}_i$ vectors can be seen as linear classifiers in the target domain and the matrix $\mat{R}$ combines all of them taking the dependencies between classes into account. This is an interesting and important aspect of our method in scenarios with a large number of categories. A linear classifier $\infosingle_i$ is mapped to the target domain by:
\begin{align}
  \tilde{\infosingle}_i &= \datransform^T \infosingle = \sum\limits_{i'=1}^m \vec{\beta}_i \, \rho_{i,i'}
\end{align}
and therefore uses correlations to other categories, which is similar to transfer learning approaches~\cite{Tommasi09:myk}. 
To allow for efficient $\alpha$-updates in \equationname~\eqref{eq:alphaupdate}, we further need to consider an efficient calculation of the feature vector norm $\|\vec{d}_\ell\|^2$:
\begin{align}
\|\vec{d}_\ell\|^2 &= \| \infosingle_i \cdot \dainputsingle_j^T \|^2 = \| \infosingle_i \|^2 \cdot \| \dainputsingle_j \|^2 \enspace.
\end{align}
Finally the update formula in \equationname~\eqref{eq:wupdate} can be translated into updating $\vec{\beta}_i$ in only $\landau(\tilde{\dimension})$ operations:
\begin{align}
  \vec{\beta}_i \leftarrow \vec{\beta}_i + \triangle \alpha_\ell \, t_\ell \, \dainputsingle_j \enspace.
\end{align}

\begin{table}
  \centering
    \small
  \begin{tabular}{ccc}
    \toprule
                     & $\alpha_\ell$ update & $\vec{W}$ update \\
    \midrule
    \textbf{Our approach} & $\mathcal{O}(\tilde{\dimension})$ & $\mathcal{O}( \tilde{\dimension} )$\\ 
    Direct rep. of $\datransform$ & $\mathcal{O}(\dimension \cdot \tilde{\dimension})$ & $\mathcal{O}(\dimension \cdot \tilde{\dimension})$\\ 
    \midrule
    Bregman opt. (kernel)~\cite{Kulis11:WYS} & - & $\mathcal{O}(\noe \cdot \tilde{\noe})$ \\
    Bregman opt. (linear) & - & $\mathcal{O}(\dimension \cdot \tilde{\dimension})$ \\
    \bottomrule
  \end{tabular}
  \caption{Asymptotic times for one iteration of the optimization, where a single constraint is taken into account. There are $\noe$ source training points of dimension $\dimension$ and $\tilde{\noe}$ target training points of dimension $\tilde{\dimension}$.}
  \label{tab:asymptotictimes}
\end{table}

\subsection{Algorithmic details and complexity}
\label{sec:algodetails}

  \begin{figure}
    \small
    \rule{\linewidth}{1pt}
    \vspace{-20pt}
    \begin{center}
    Optimization of $\datransform$ in our method
    \end{center}
    \vspace{-14pt}
    \rule{\linewidth}{0.5pt}\vspace{-2pt}
      \begin{enumerate}
      \item For $1 \leq i,i' \leq K$: $\rho_{i,i'} = \infosingle_i^T \infosingle_{i'}$ 
      \item For $1 \leq j \leq \tilde{\noe}$: $q_j = \| \dainputsingle_j \|^2$
      \item Repeat until convergence of $\vec{\alpha}$
        \begin{enumerate}
          \item Loop through the active set $\ell = (i,j) \in \mathcal{S}$
          \begin{enumerate}
            \item $s = \sum_{i'=1}^m \rho_{i,i'} \, \vec{\beta}_{i'}^T \dainputsingle_j$ using cached $\vec{\beta}_{i'}^T \dainputsingle_j$
            \item $G = \delta\left[ \dalabelsingle_j = i \right] \cdot s + \lambda \alpha_\ell - 1$
            \item $PG = \begin{cases} 
                              G & \alpha_\ell > 0\\
                              \min(G,0) & \alpha_\ell = 0\\
                        \end{cases}$
            \item if $PG \neq 0$
              \begin{enumerate}
                \item $\alpha_\ell \leftarrow \max( \alpha_\ell - G/(q_j \cdot \rho_{i,i} + \lambda ),0)$
                \item $\vec{\beta}_i \leftarrow \vec{\beta}_i + \triangle \alpha_\ell \, \delta\left[ \dalabelsingle_j = i \right] \, \dainputsingle_j$
              \end{enumerate}
          \end{enumerate}
        \end{enumerate}
      \end{enumerate}
    \vspace{-8pt}
    \rule{\linewidth}{1pt}
    \vspace{-.2cm}
    \caption{Pseudo code for $\datransform$ optimization without shrinking heuristics and caching details.}
    \label{fig:algo}
  \end{figure}

  In this section, we briefly discuss some implementation details of the solver used in our experiments (\sectionname~\ref{sec:experiments}).
  Code for our efficient dual coordinate descent transform solver, adapted from \texttt{liblinear}~\cite{Fan08:LLL}, will be made publicly available online.
     The shrinking heuristics presented in \cite{Hsieh08:DCD} that maintain a set $\mathcal{S}$ of dual variables that have been set to zero during optimization and that are likely not to change in the future are also implemented in our approach. 
    An algorithmic outline of our approach is given in \figurename~\ref{fig:algo}.

  \myparagraph{Computational complexity}
    The asymptotic times are summarized in \tablename~\ref{tab:asymptotictimes}. While the asymptotic time for the kernel Bregman optimization used in \cite{Kulis11:WYS,Hoffman13:ELD} depends on the number of source examples, the time we need to iteratively take one constraint into account is independent of the number of examples in either the source or target domain. One pass over all constraints takes time $\mathcal{O}(\tilde{n} \cdot m)$, which finally leads to a linear asymptotic time in the product of the number of target points and the target dimension, independent of the size of the source training set. Therefore, our method allows for using transform-based adaptation in large-scale settings, where previous approaches~\cite{Kulis11:WYS,Saenko10:AVC} were unable to run at all.

  \myparagraph{Identity regularizer}
    As described in previous sections, the transformation $\datransform$ has a low-rank structure when using the original MMDT formulation. In situations with only a small number of categories, this can be too restrictive for the class of transformations. However, when using the identity regularizer $\| \datransform - \identityMatrix{} \|^2_F$,
    we obtain 
    $\datransform = \identityMatrix{} + \sum_i \infosingle_i \vec{\beta}_i^T$,
    which allows to estimate full rank matrices. The efficient updates in each coordinate descent iteration do not change significantly and are omitted here due to the lack of space.

  \myparagraph{Caching techniques}
    As mentioned earlier, we cache the scalar products $\vec{\beta}_i^T \dainputsingle_j$ to allow for fast computation. Each time the vector $\vec{\beta}_i$ is updated, all $\tilde{\noe}$ cached values $\vec{\beta}_i^T \dainputsingle_j$ are invalid and have to be updated in one of the next steps where $\dainputsingle_j$ is taken into account. When using a fully randomized order of the dual variables $\alpha_\ell$ as suggested by \cite{Hsieh08:DCD}, this invalidation happens on average every $K$th step leading to a low probability that the cached value can be used in between. For this reason, we only consider a random order of $j$ and iterate normally through all the $K$ categories. Therefore, we can use the cached values in each of the $K$ blocks.

  \myparagraph{Convergence properties}
    Our solver maintains all the convergence properties of dual coordinate descent solvers. In particular, we have at least a linear convergence rate~\cite[Theorem 1]{Hsieh08:DCD} and an $\epsilon$-accurate solution can be obtained in $\landau(- \log(\epsilon))$ iterations.

\section{Domain adaptation datasets}
  \label{sec:imagenetdataset}

  In the following, we briefly describe the datasets used in our experiments for the source as well as the target domain.

  \myparagraph{ImageNet ILSVRC2010 to SUN2012}

    Whereas ImageNet images were obtained using object category names and therefore contain a large portion of advertisement images, the creation of the SUN database was done by searching for scene categories and labeling objects in the images afterwards. Therefore, there is a significant domain shift between the two datasets (\figurename~\ref{fig:bias}).
    In fact, Torralba and Efros's experiments in~\cite{ref:Efros-dataset-bias-cvpr2011} consistently showed that the domain shift between ImageNet and SUN is one of the most severe among all pairs of benchmark datasets they surveyed.

    For this reason, we assembled a new challenge for domain adaptation methods by matching a subset of the object categories from the SUN2012 dataset~\cite{Xiao10:sun} (target domain) with the ones present in the hierarchy of the ImageNet 2010 challenge~\cite{Berg10:Lsv} (source domain). The matching of the category names in both datasets is done by using the manually maintained WordNet matchings of the SUN2012 dataset~\cite{Xiao10:sun}. Using the WordNet descriptions, a large set of SUN2012 descriptions can be mapped to nodes of the WordNet subgraph related to the ILSVRC2010 challenge; \ie, to sets of ILSVRC2010 categories (leaf nodes). Finally, we consider pairs of SUN2012 labels and ILSVRC2010 category sets that lead to more than $20$ examples. This leads to a total of $84$ categories\footnote{tree, chair, cabinet, table, lamp, curtain, box, car, bed, mountain, desk, fence, mirror, skyscraper, bottle, rug, basket, bench, towel, vase, bannister, ball, stove, bookcase, magazine, refrigerator, bucket, clock, glass, hat, oven, boat, fan, shoe, dishwasher, telephone, airplane, loudspeaker, apparel, keyboard, bar, gate, bus, mug, bridge, umbrella, bicycle, backpack, laptop, washer, bathtub, roof, pitcher, fish, tower, flower, apple, file, teapot, minibike, printer, garage, guitar, ashcan, dog, dune, piano, ship, crane, newspaper, mouse, microphone, cliff, bell, elephant, shirt, toaster, orange, remote control, knife, helmet, grape, stick, shop}. The final set of examples consists of cropped bounding boxes not labeled as difficult or truncated. Classification with these examples without context knowledge can be considered as very challenging.

    To allow for easy reproducibility of the results, we use the bag of visual words (BoW) features provided for the ImageNet challenge. Furthermore, features in the SUN database are extracted by computing bag of visual words features inside of the given bounding boxes. This is also done with the feature extraction code provided for the ImageNet challenge.

  \myparagraph{Bing/Caltech256 dataset}

  We also use the Bing dataset of \cite{BergamoTorresani10}, which contains images for each category of the Caltech256 dataset. In contrast to the ImageNet/SUN2012 scenario, both datasets have been created using internet search images and category keywords. In total, this dataset consists of 256 object categories. Features for this dataset are provided by the authors of \cite{BergamoTorresani10}.

\section{Experiments}
\label{sec:experiments}

  In our experiments, we give empirical validation for the following claims:
  \vspace{-2pt}
  \begin{enumerate}
    \itemsep-2pt
    \item Our optimization algorithm allows for significantly faster learning than the one used by \cite{Hoffman13:ELD} without loss in recognition performance (\sectionname~\ref{sec:mediumscale}).
    \item Our transform-based approach can be used for large-scale domain adaptation datasets and achieves state-of-the-art performance, significantly outperforming the geodesic flow kernel method of \cite{ref:gong12_gfk} (\sectionname~\ref{sec:large}).
    \item We can learn a transformation between large-scale datasets that can be used for transferring new category models without any target training examples (\sectionname~\ref{sec:newcategories}) even in the case of different feature dimensions (\sectionname~\ref{sec:diffdim}).
  \end{enumerate}

  \subsection{Baseline methods}
    We compare our approach to the standard domain adaptation baseline, which is a linear SVM trained with only target or only source training examples (\emph{SVM-Target}/\emph{SVM-Source}). 
    Note that for new category experiments, where some classes do not have training examples in the target domain, the \emph{SVM-Target} baseline cannot be used.
    Furthermore, we evaluate the performance of the geodesic flow kernel (\emph{GFK}) presented by \cite{ref:gong12_gfk} and integrated in a nearest neighbor approach.
    The metric learning approach of \cite{Kulis11:WYS} (\emph{ARC-t}) and the shared latent space method of \cite{ref:duan_icml12} (\emph{HFA}) can only be compared to our approach
    in a medium-scale experiment which is tractable for kernelized methods. For our experiments, we always use the source code from the authors.

    We refer to our method as large-scale max-margin domain transform (\emph{LS-MMDT}) in the following.

  \subsection{Comparison to other adaptation methods}
  \label{sec:mediumscale}
    
    \begin{figure}
      \centering
      \includegraphics[angle=270,width=0.49\linewidth]{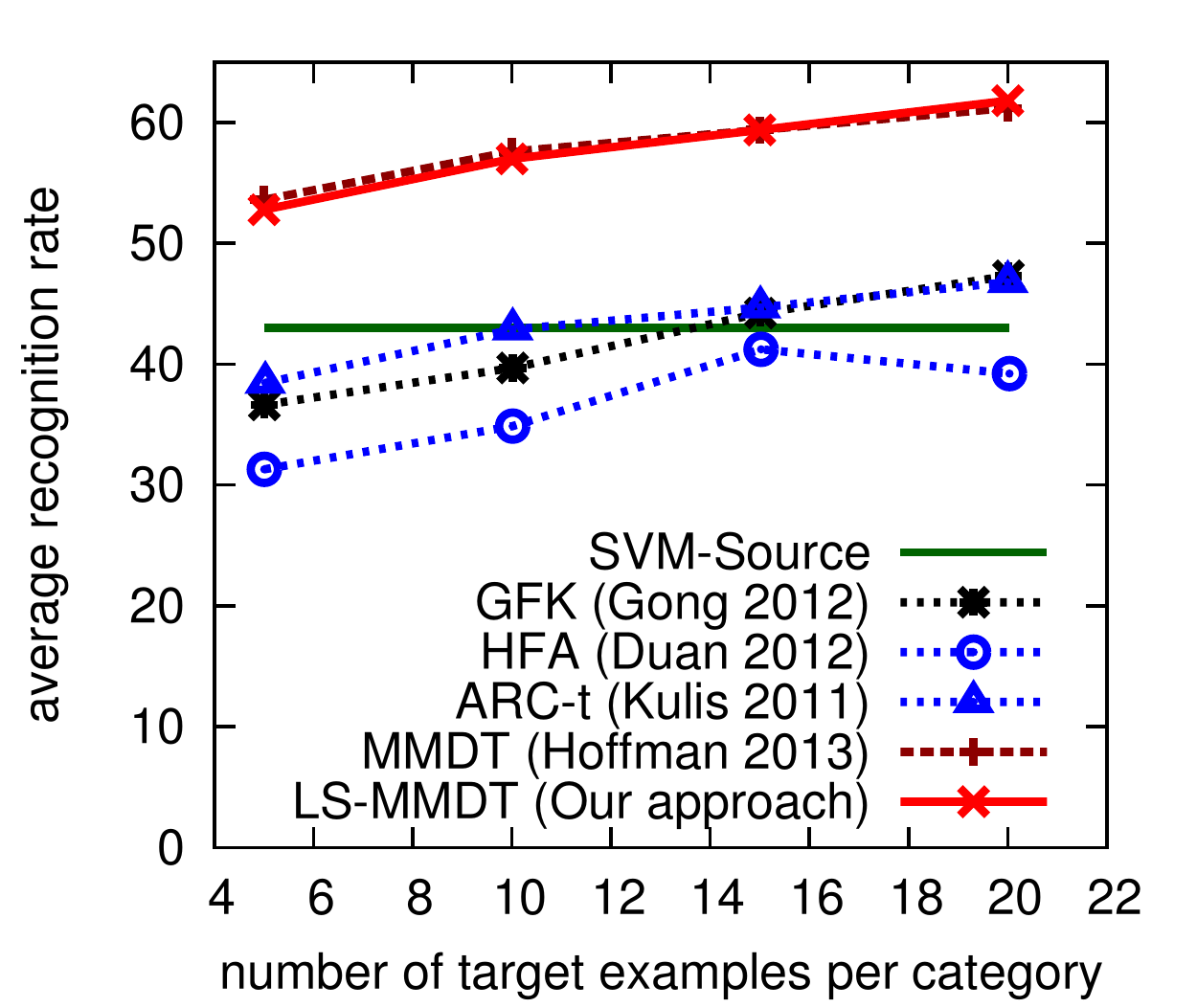}
      \includegraphics[angle=270,width=0.49\linewidth]{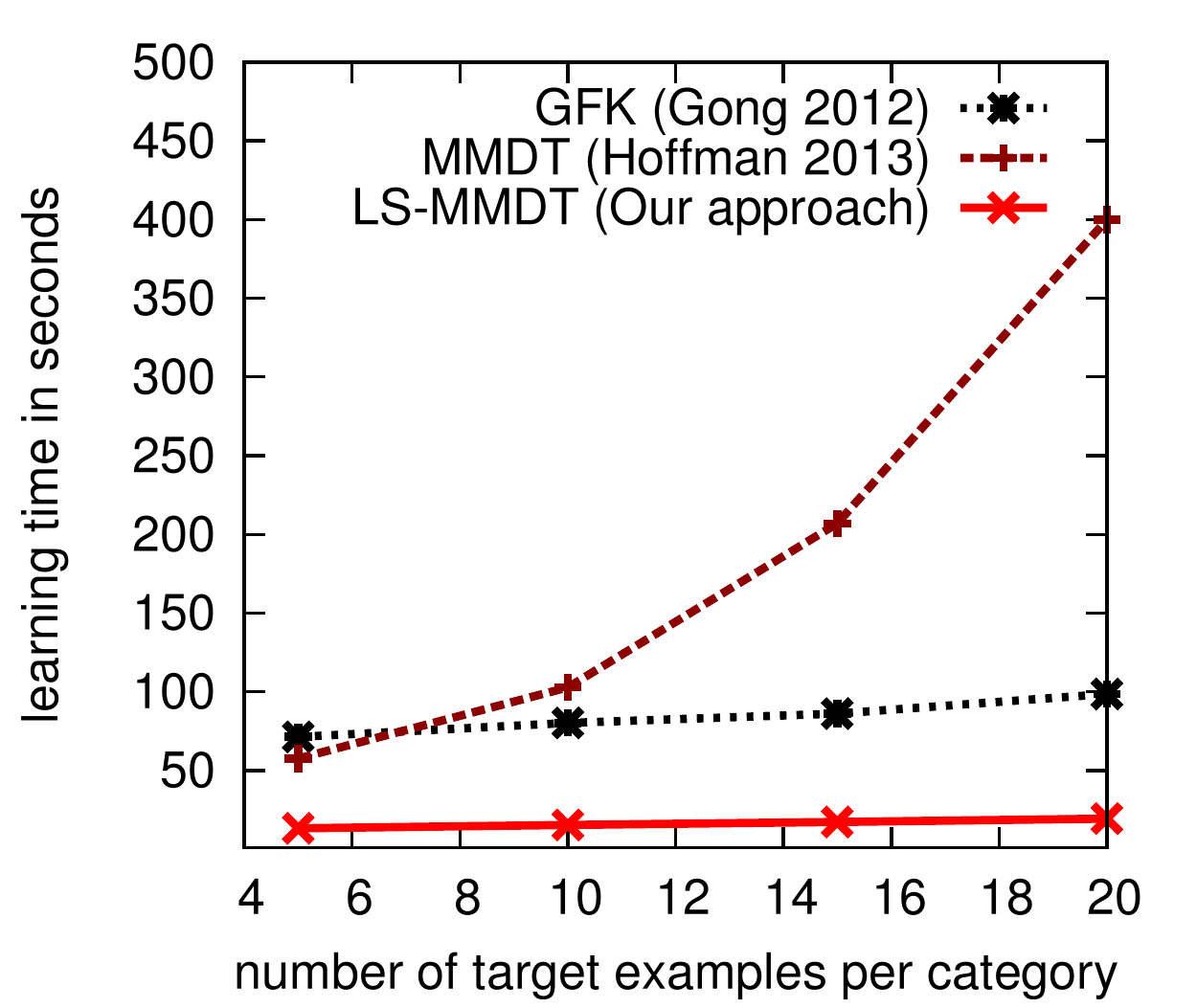}
      \caption{Medium-scale experiment: recognition rates and learning times when using the first 20 categories of the Bing/Caltech256 (source/target) dataset. Times of ARC-t~\cite{Kulis11:WYS} and HFA~\cite{ref:duan_icml12} are off-scale (12min and 55min for 10 target points per category).}
      \label{fig:smallscale}
    \end{figure}

    We first evaluate our approach on a medium-scale dataset comprised of the first 20 categories of the Bing/Caltech dataset. This setup is also used in \cite{Hoffman13:ELD} and allows us to compare our new optimization technique with the one used by \cite{Hoffman13:ELD} and also with other state-of-the-art domain adaptation methods~\cite{Kulis11:WYS,ref:duan_icml12,ref:gong12_gfk}. We use the data splits provided by \cite{BergamoTorresani10} and the Bing dataset is used as source domain with $50$ source examples per category. \figurename~\ref{fig:smallscale} contains a plot for the recognition results (left) and the training time (right plot) with respect to the number of target training examples per category in the Caltech dataset. 
    As \figurename~\ref{fig:smallscale} shows, our solver is significantly faster than the one used in \cite{Hoffman13:ELD} and achieves the same recognition accuracy. Furthermore, it outperforms other state-of-the-art methods, like ARC-t~\cite{Kulis11:WYS}, HFA~\cite{ref:duan_icml12}, and GFK~\cite{ref:gong12_gfk}, in both learning time and recognition accuracy.

  \subsection{Experiments with a large number of categories}
  \label{sec:large}

    \begin{figure}
      \centering
      \includegraphics[angle=270,width=0.49\linewidth]{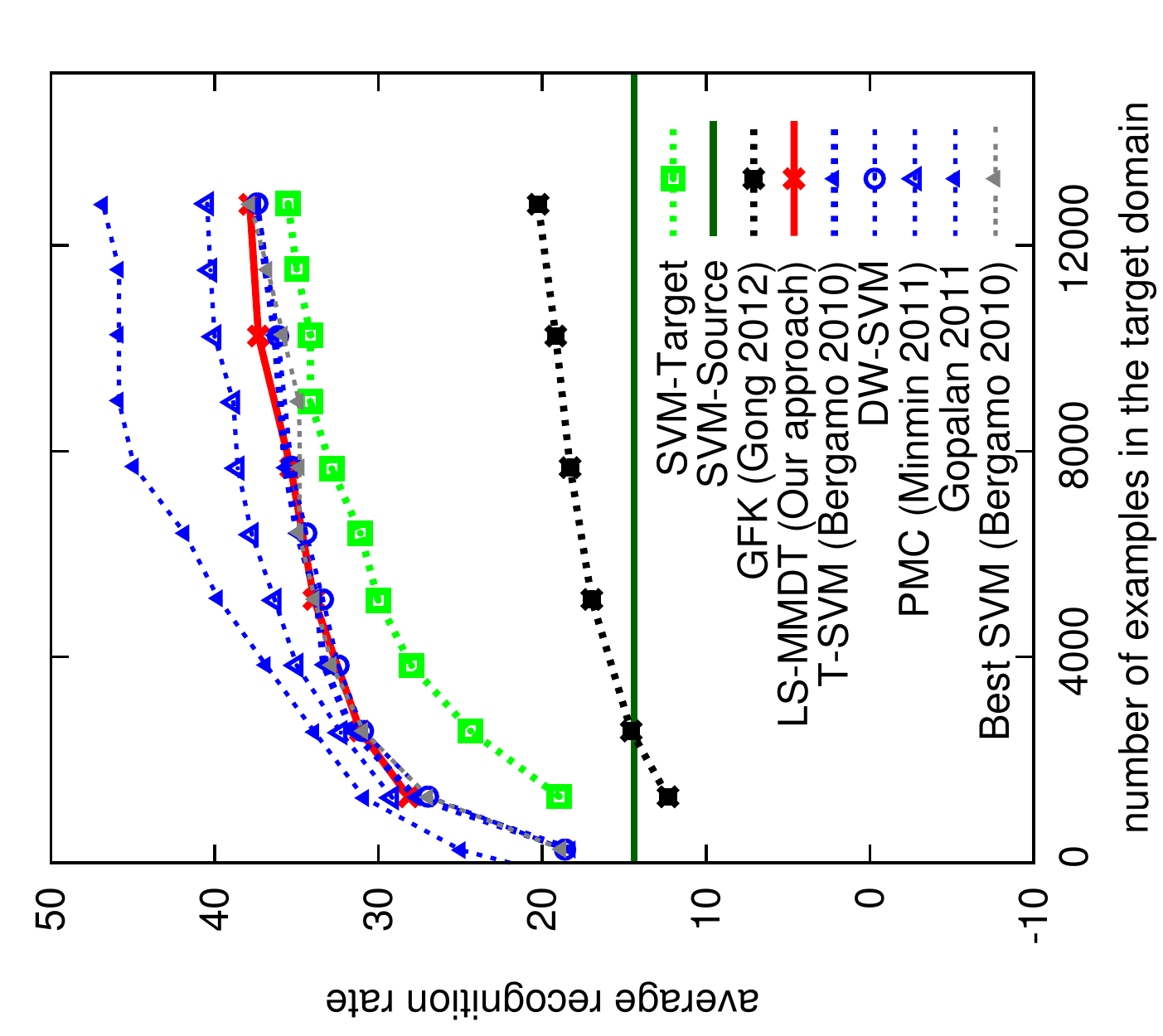}
      \includegraphics[angle=270,width=0.49\linewidth]{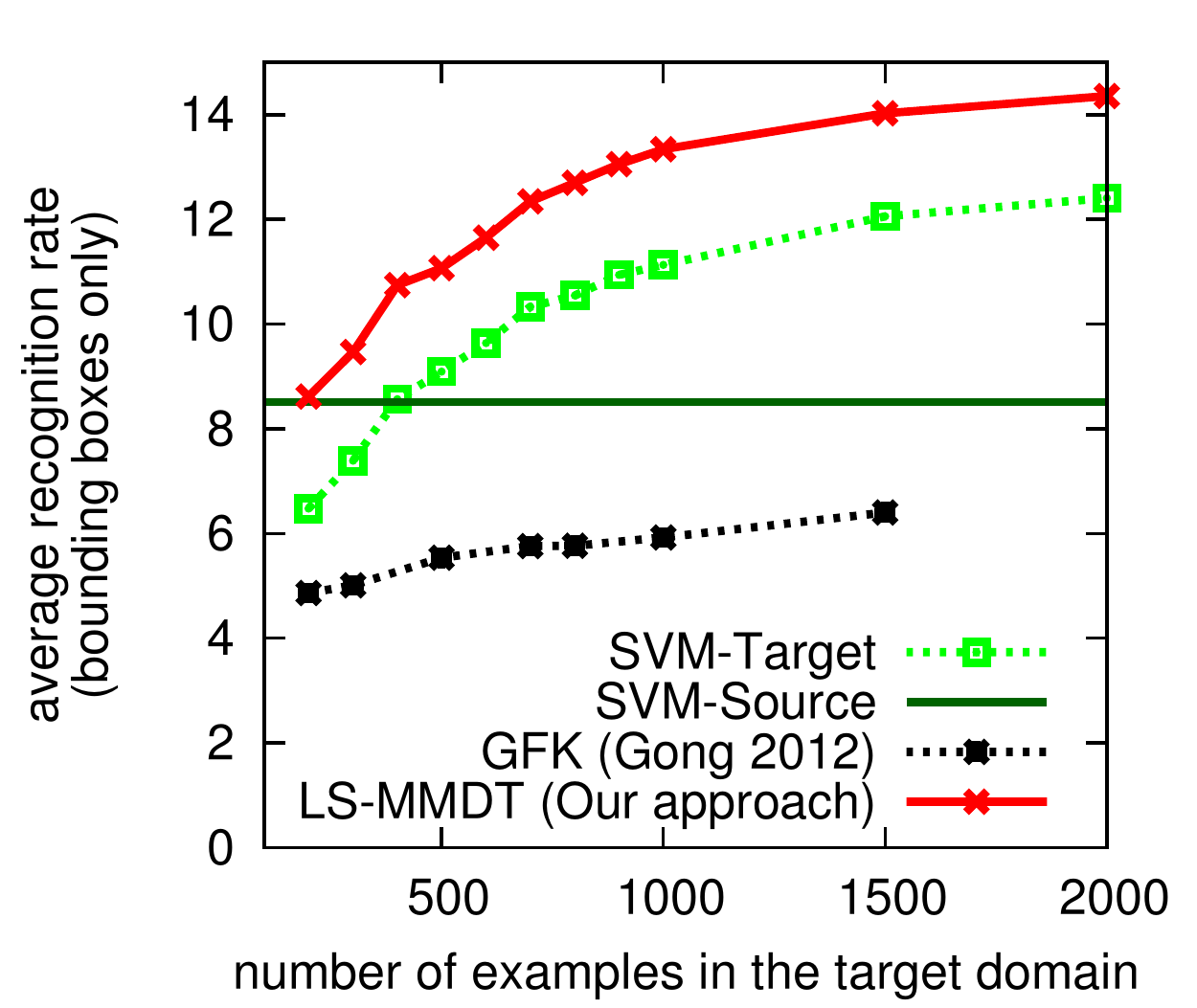}
      \caption{Large-scale experiment with the Bing/Caltech256 domain shift (76K source examples; left) as well as the ImageNet/SUN2012 domain shift (8K source examples; right) and a varying number of target examples.} 
      \label{fig:largescale}
    \end{figure}
   
    In the next experiment, we use the Bing/Caltech256 dataset~\cite{BergamoTorresani10} with all 256 categories and our Imagenet/SUN2012 subset, settings in which the optimization techniques used in \cite{Hoffman13:ELD} \emph{cannot be applied} due to the large number of target training examples. Furthermore, we test the performance of our method on the new domain dataset presented in \sectionname~\ref{sec:imagenetdataset} and we restrict the comparison to methods that provide generalization to new categories.

    The results are given in \figurename~\ref{fig:largescale} and we see that we outperform again the geodesic flow method of \cite{ref:gong12_gfk} in both cases. Focusing on the right plot (Imagenet/SUN2012 dataset), notice that our method continues to have a performance benefit over SVM-Target even as the number of labeled target examples increases.
    This is due to the small number of training examples available for several of the categories, which is typical for real-world datasets~\cite{Salakhutdinov11:LSV}. Providing more labeled training data is only possible for some of the categories and without adaptation the recognition rates of less common classes cannot be improved.

\begin{figure*}
      \centering
      SVM-Source \hspace{0.18\linewidth} SVM-Target \hspace{0.18\linewidth} \textbf{Our approach}\\

      \includegraphics[angle=90,width=0.3\linewidth]{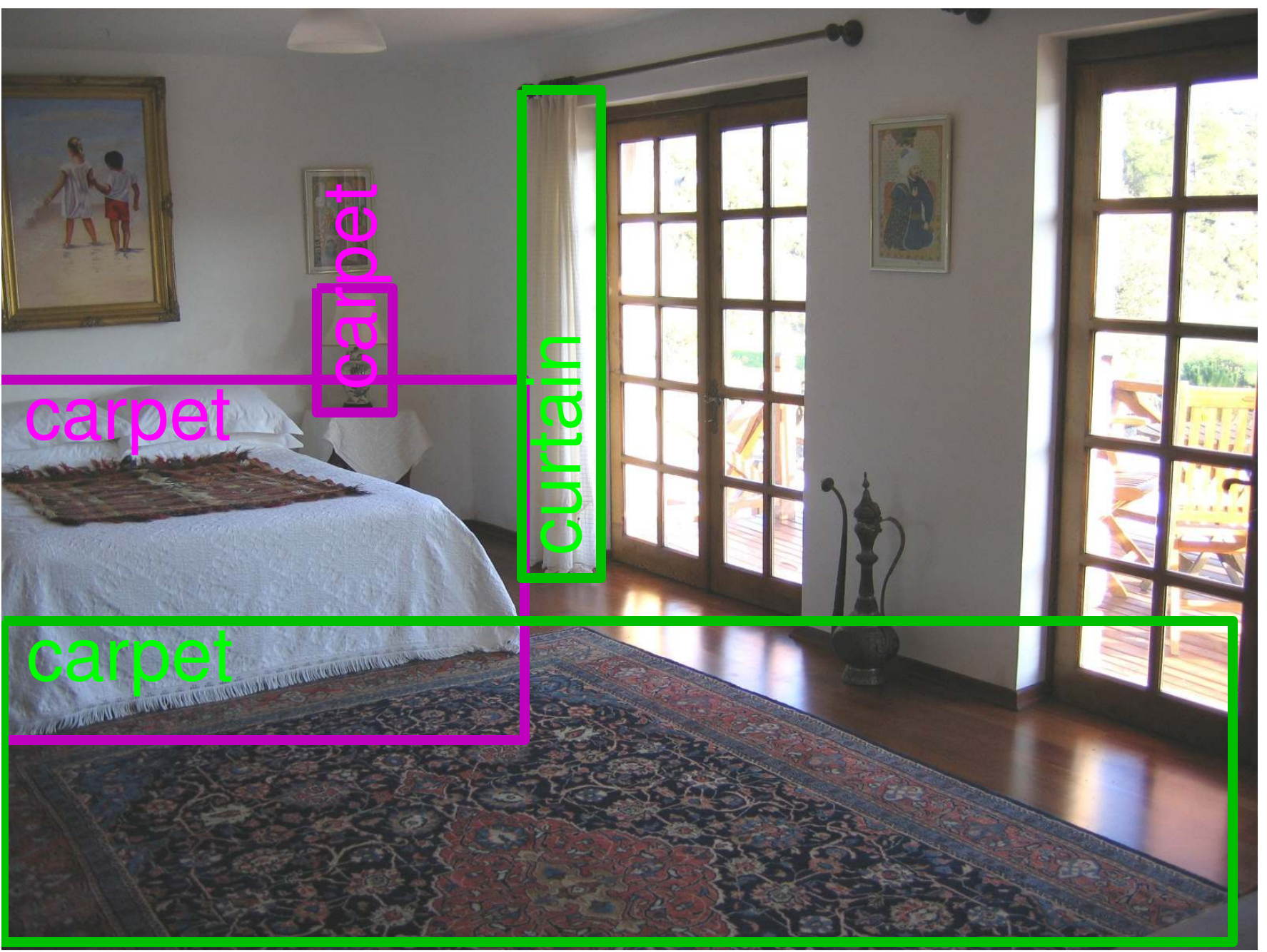}
      \includegraphics[angle=90,width=0.3\linewidth]{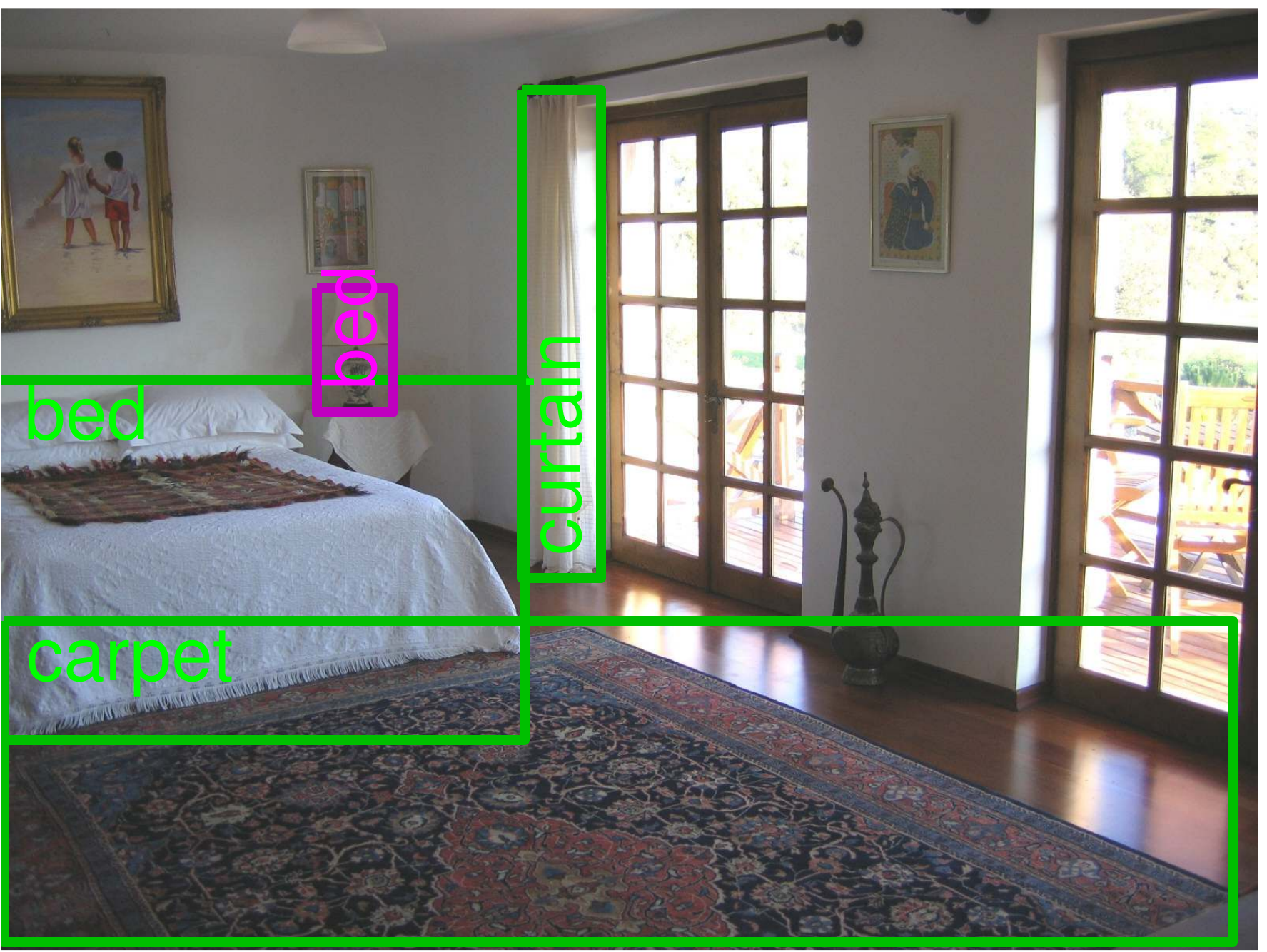}
      \includegraphics[angle=90,width=0.3\linewidth]{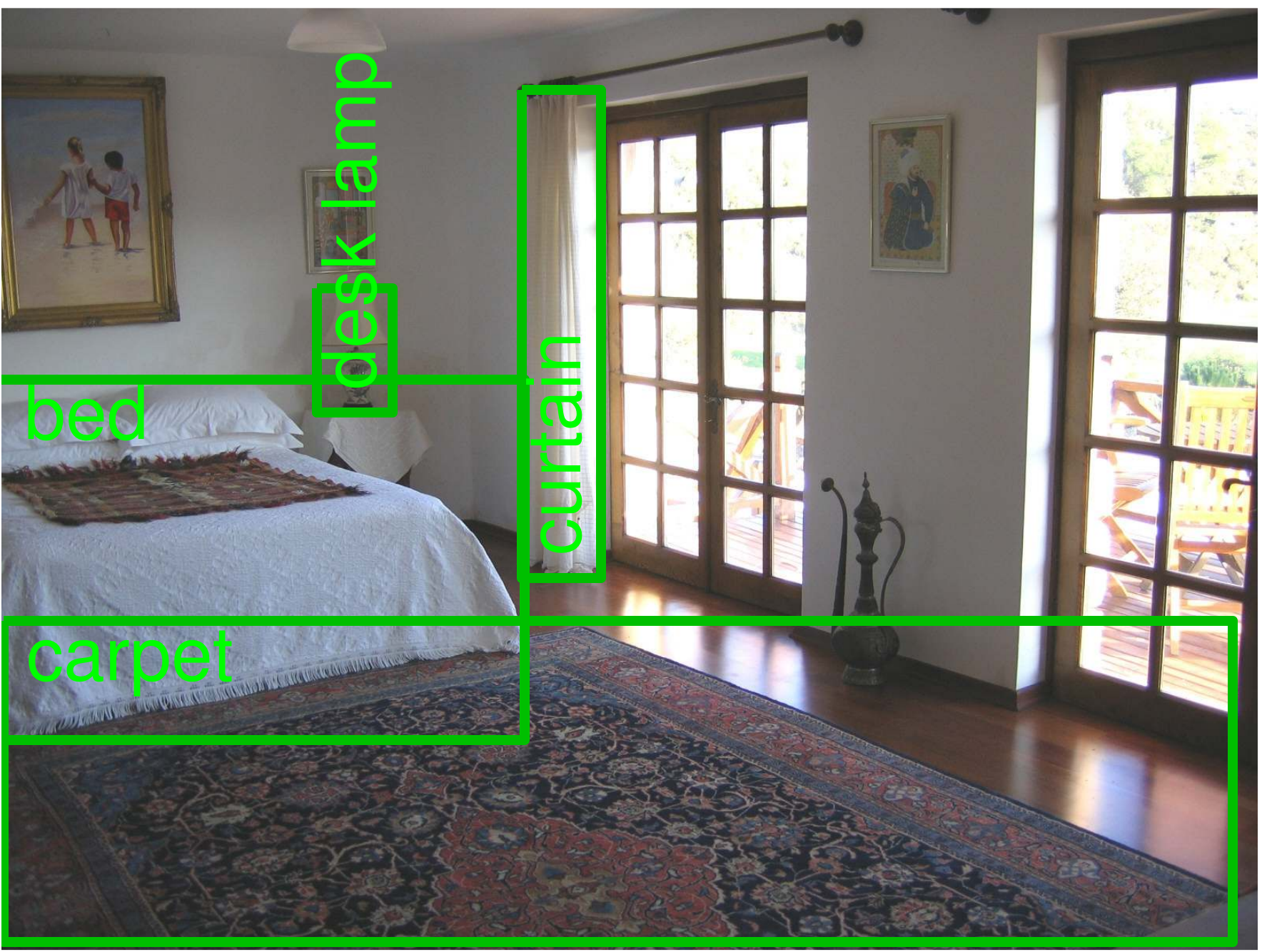}\\
      \includegraphics[width=0.3\linewidth]{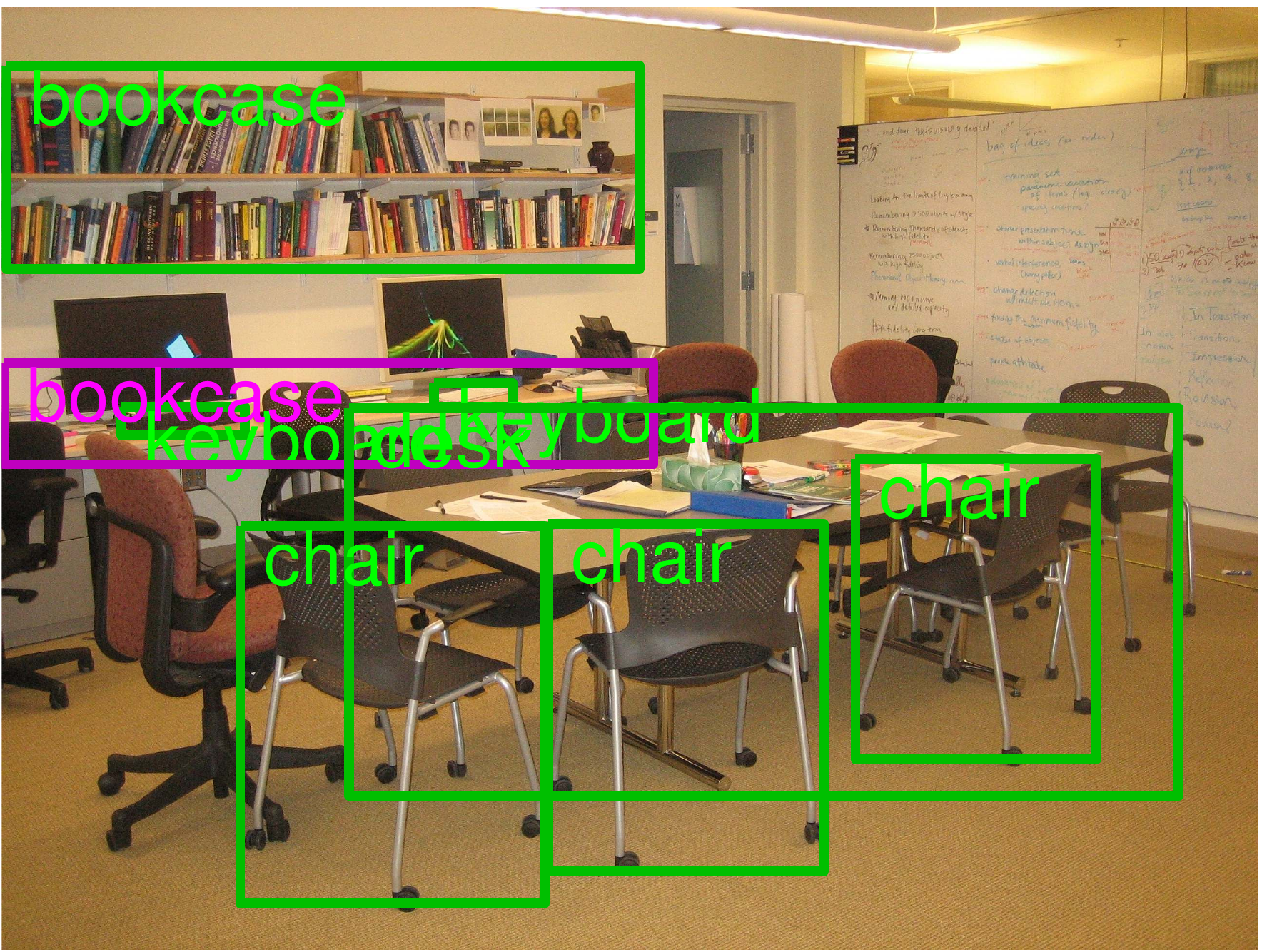}
      \includegraphics[width=0.3\linewidth]{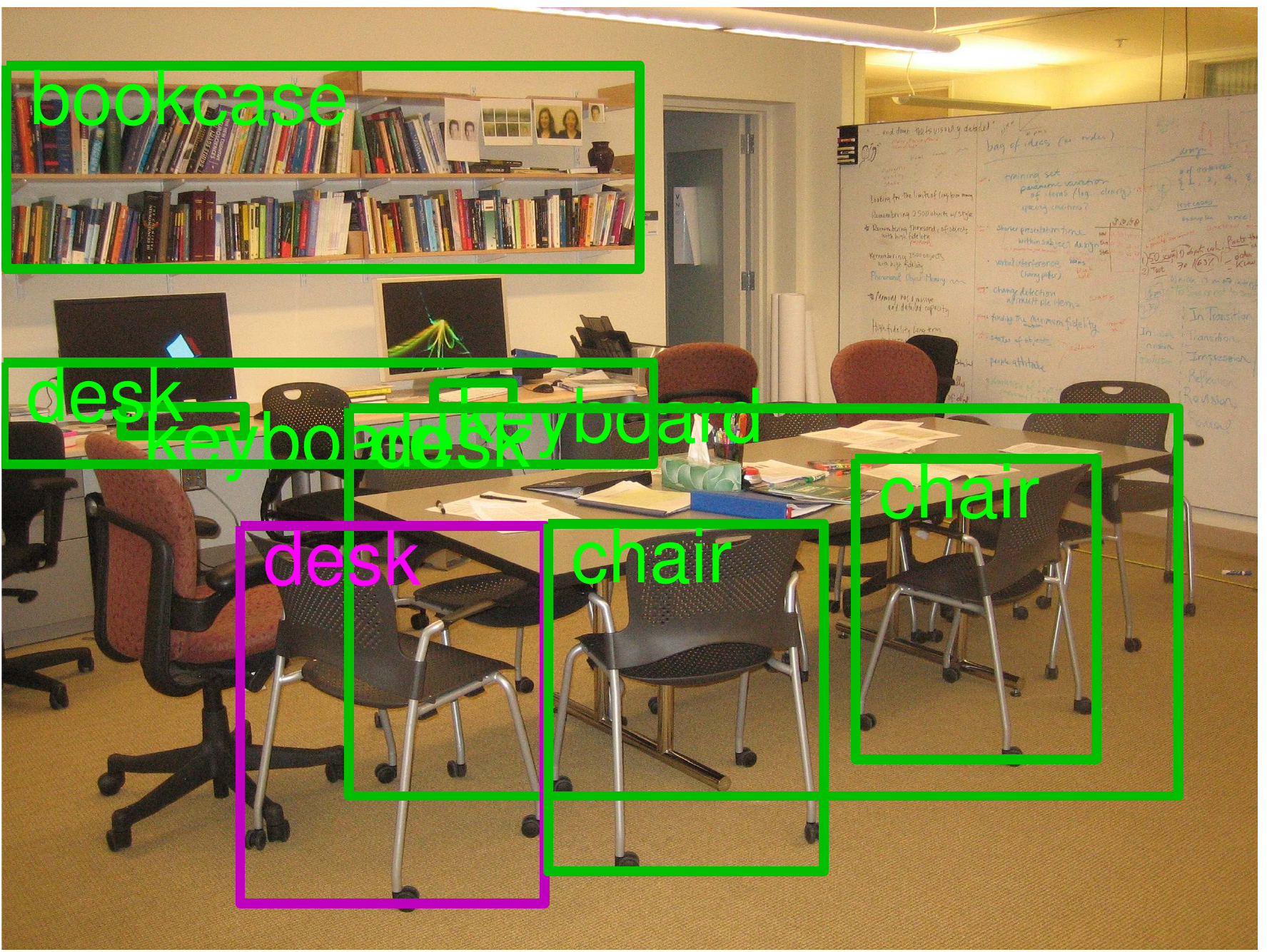}
      \includegraphics[width=0.3\linewidth]{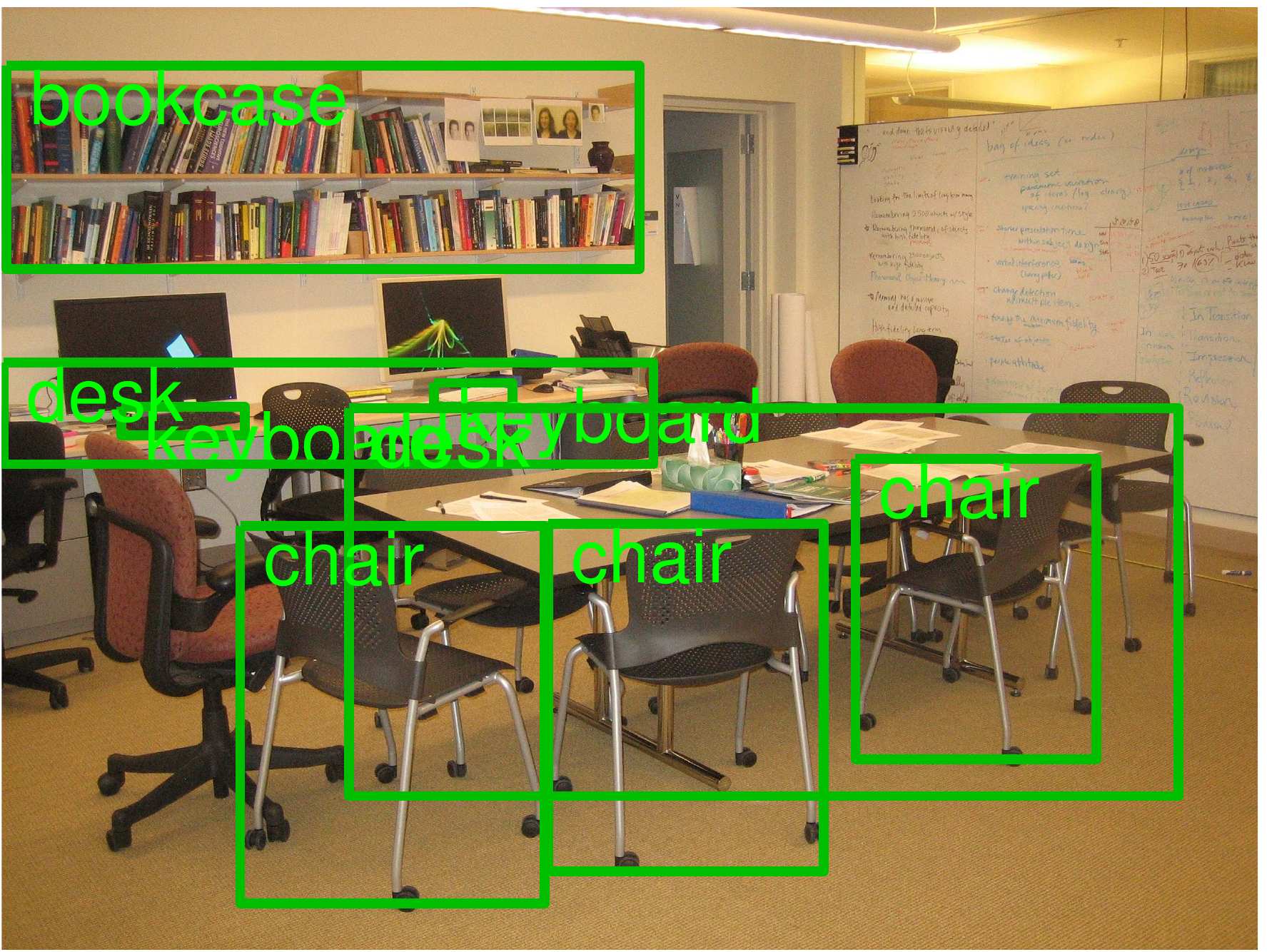}\\
      \includegraphics[width=0.3\linewidth]{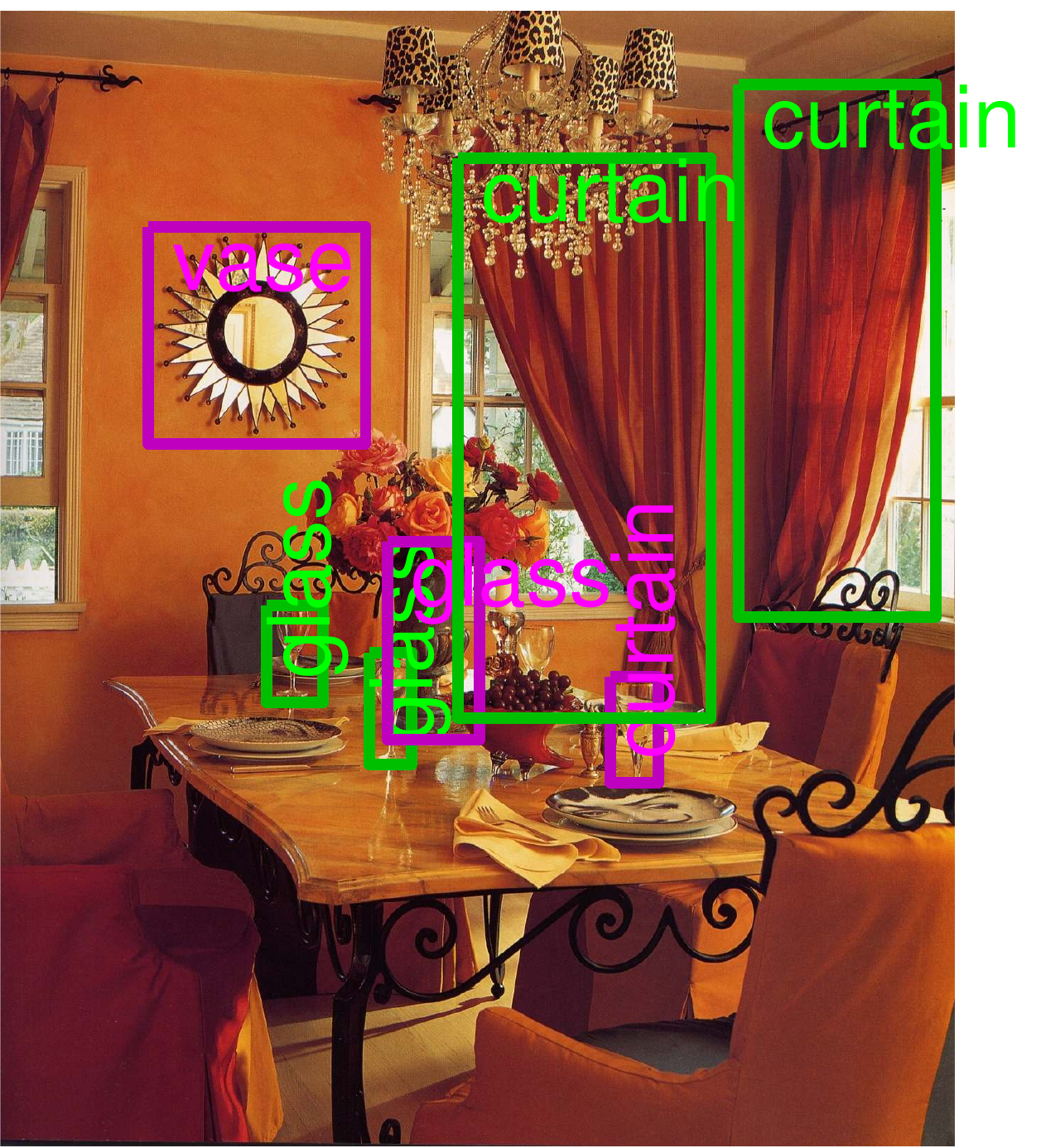}
      \includegraphics[width=0.3\linewidth]{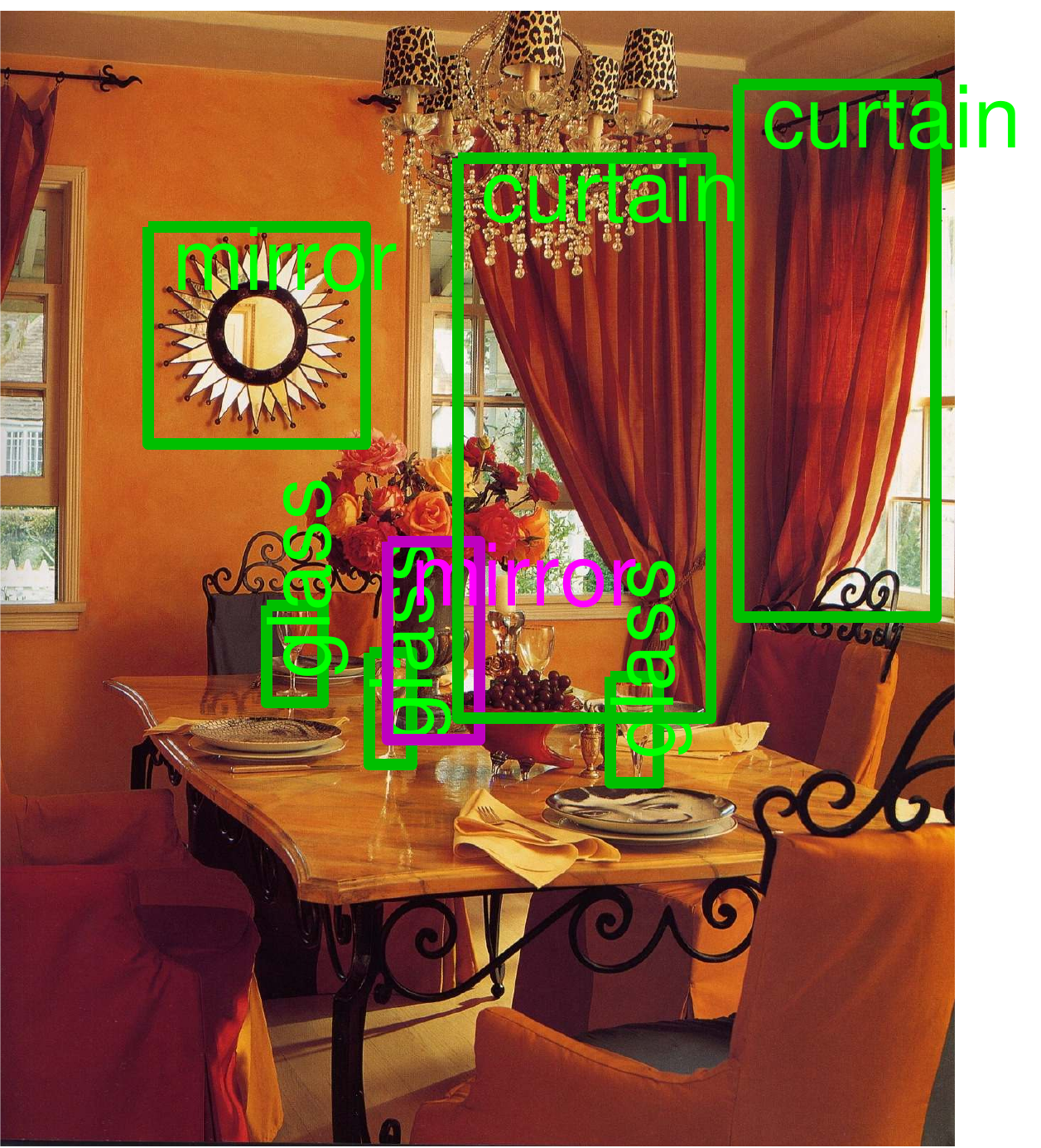}
      \includegraphics[width=0.3\linewidth]{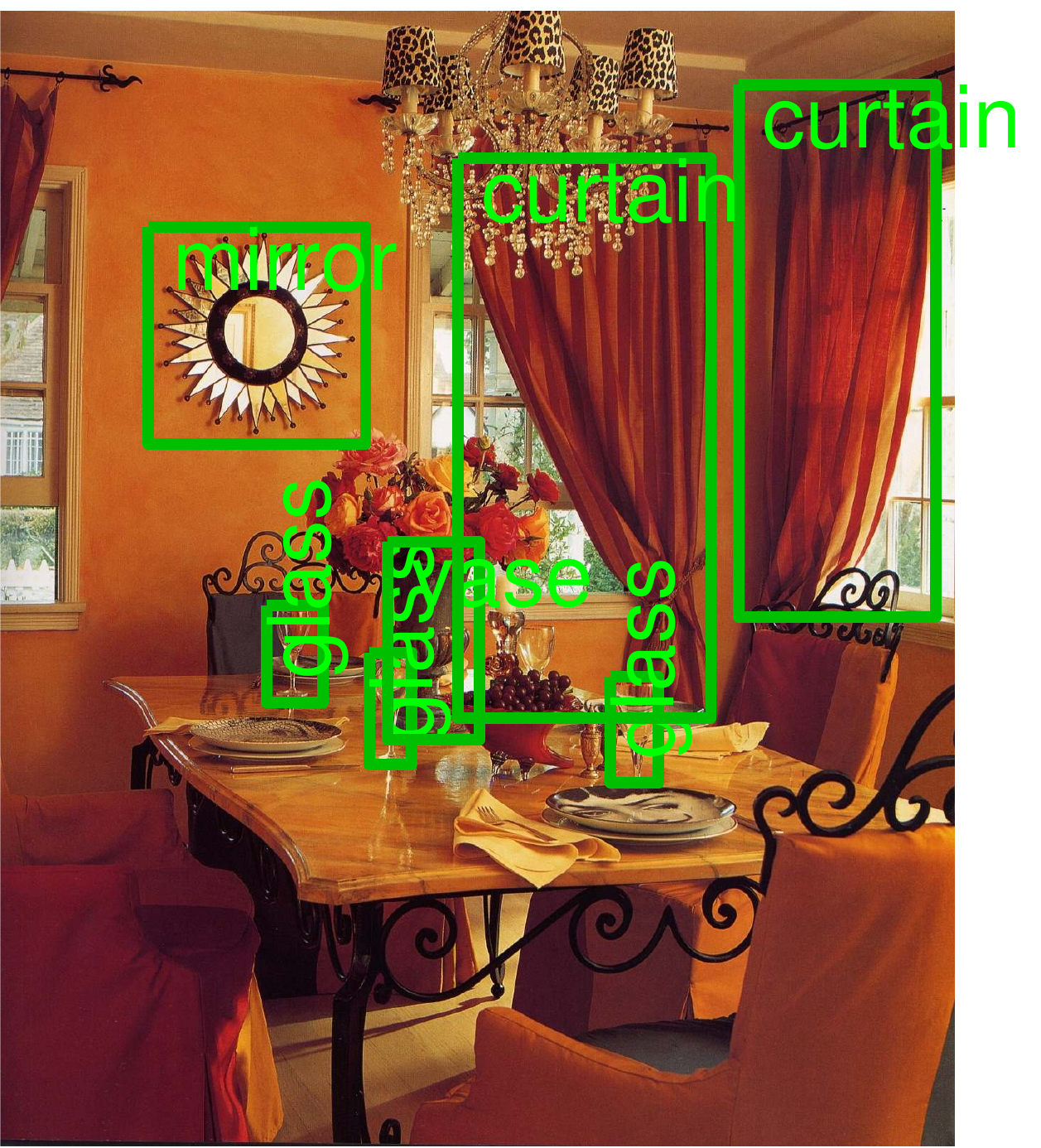}\\
      \includegraphics[width=0.3\linewidth]{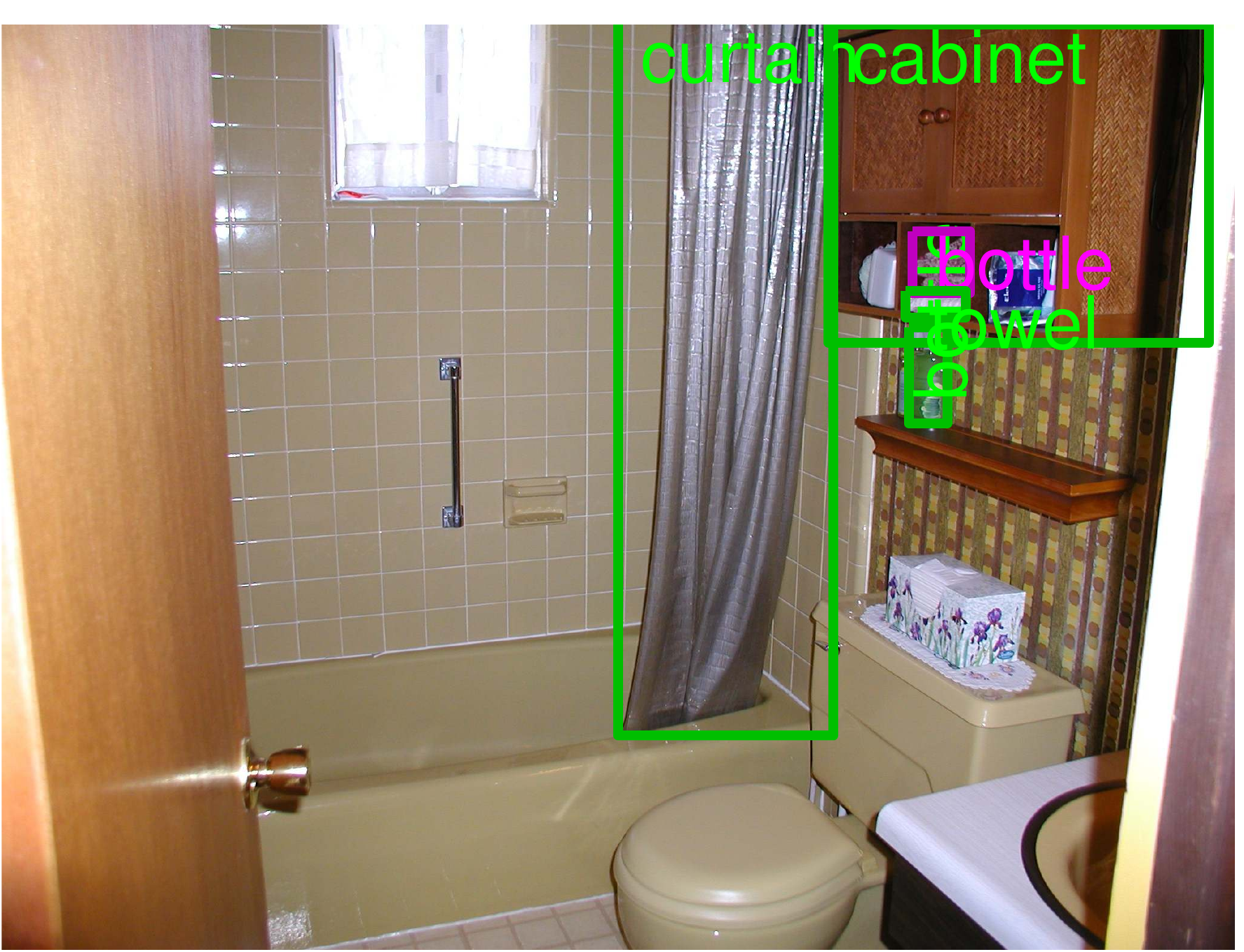}
      \includegraphics[width=0.3\linewidth]{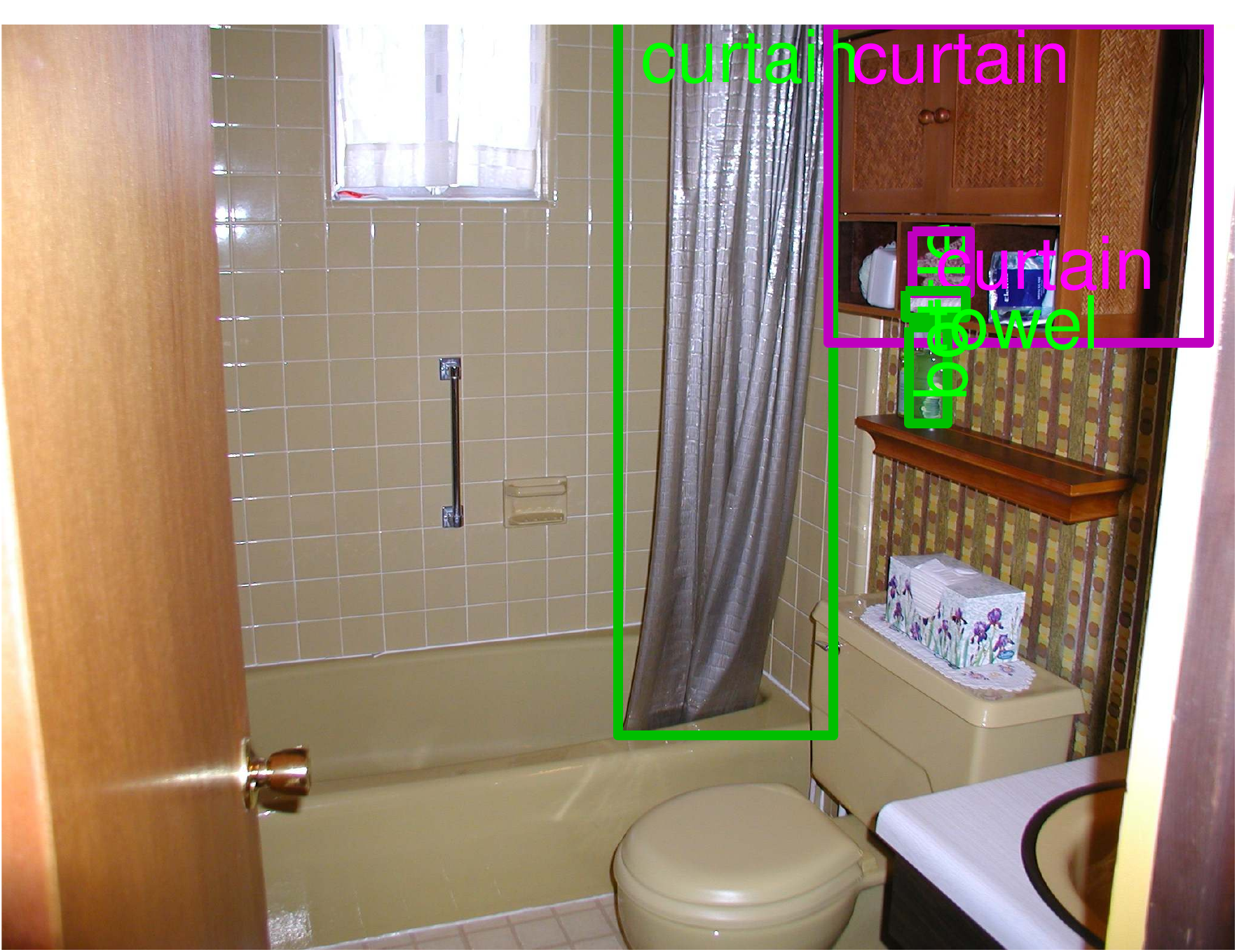}
      \includegraphics[width=0.3\linewidth]{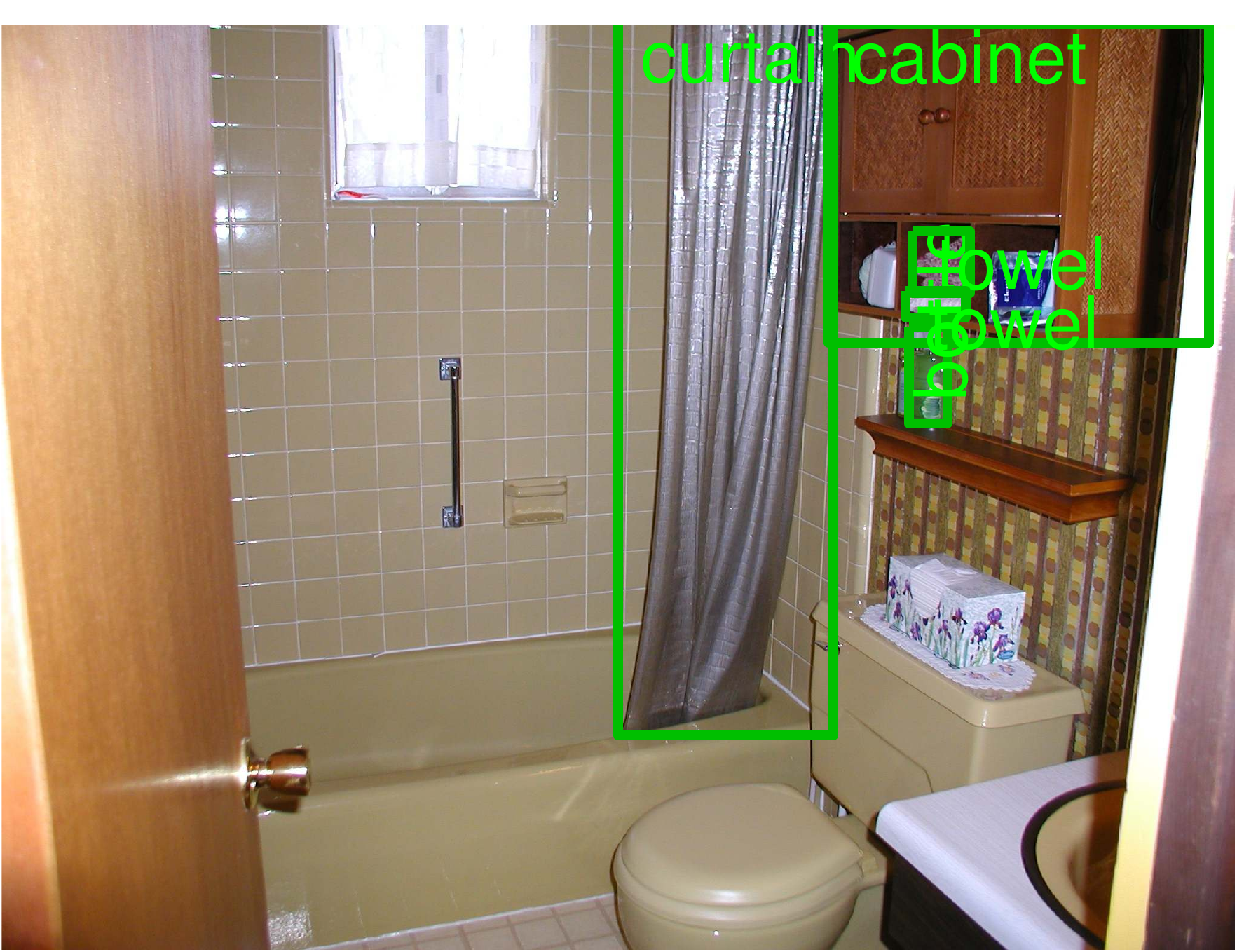}\\
      \includegraphics[width=0.3\linewidth]{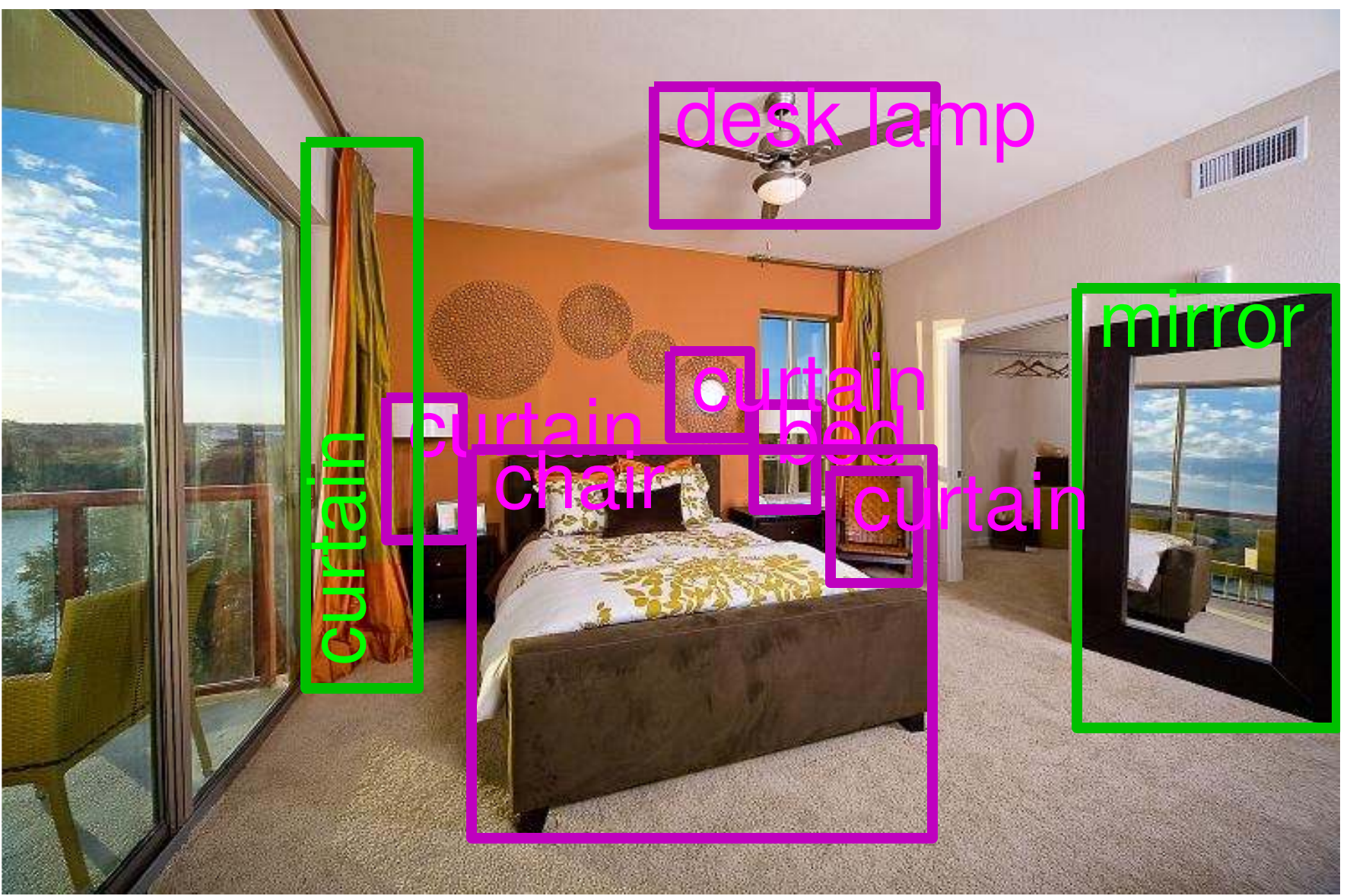}
      \includegraphics[width=0.3\linewidth]{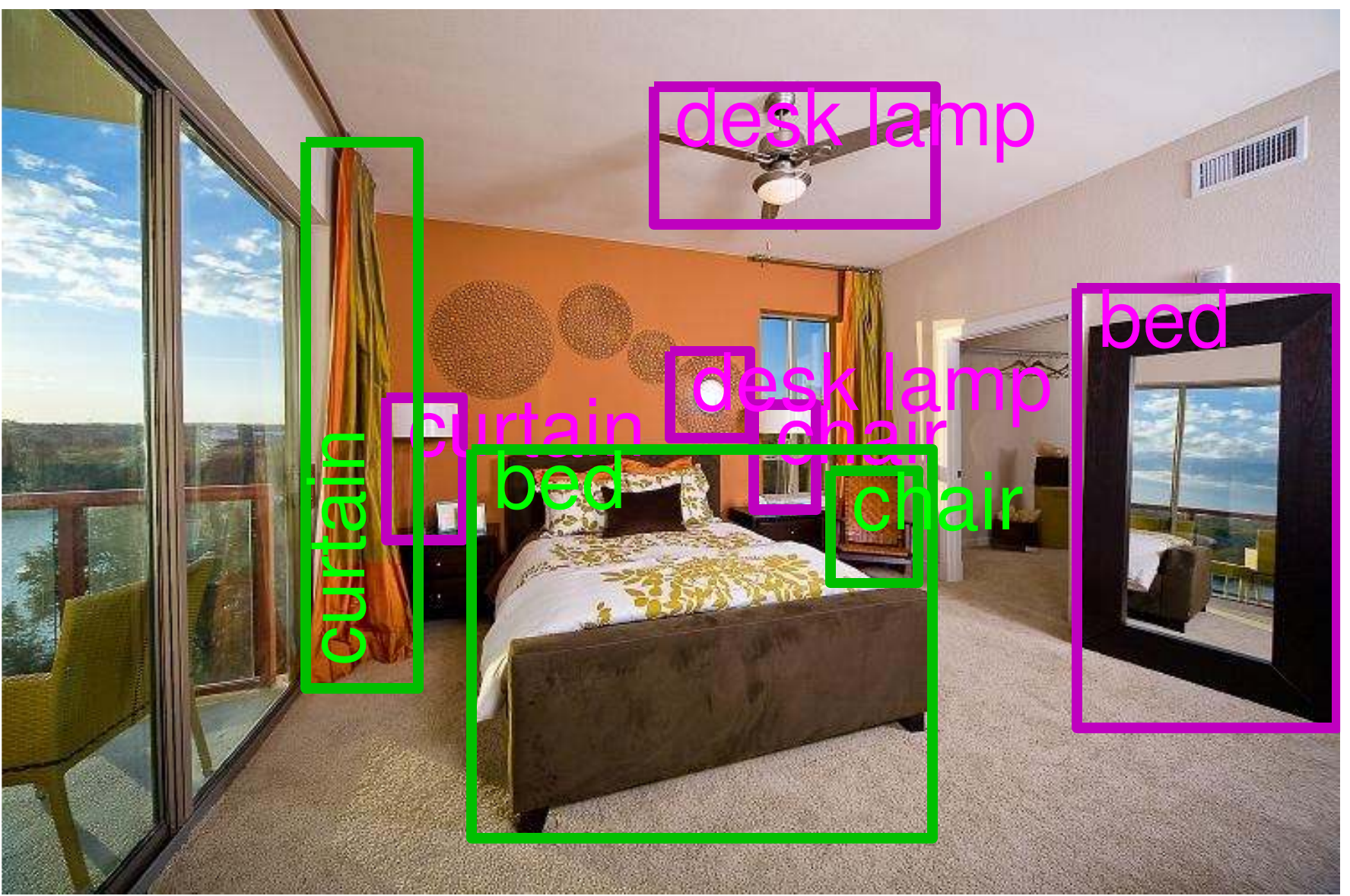}
      \includegraphics[width=0.3\linewidth]{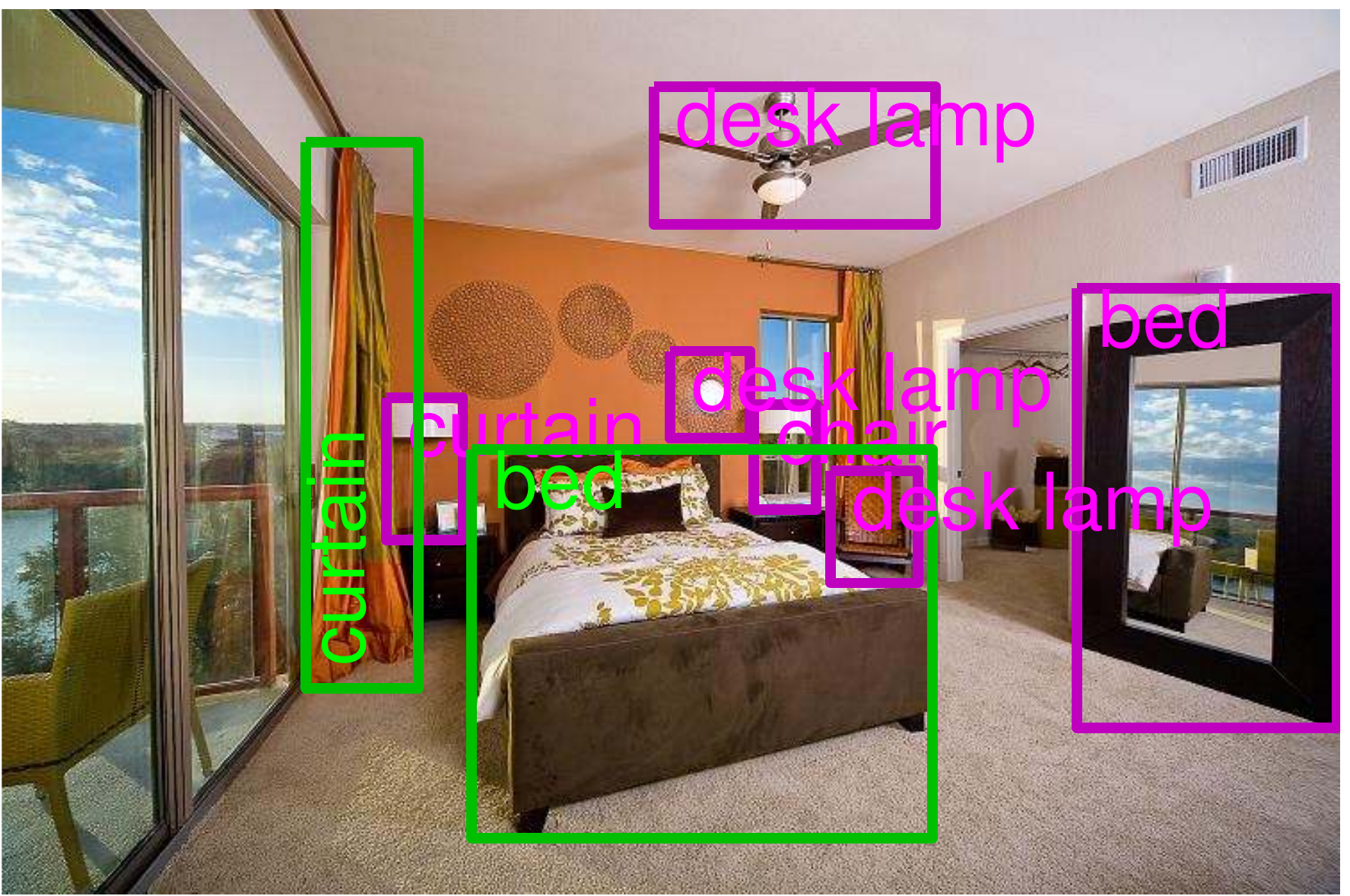}
      \caption{Results for object classification with given bounding boxes and scene prior knowledge: columns show the results of (1) SVM-Source, (2) SVM-Target, and (3) transform-based domain adaptation using our method. Correct classifications are highlighted with green borders. The figure is best viewed in color.}
      \label{fig:qualitative}
    \end{figure*}

    \figurename~\ref{fig:qualitative} shows some of the results we obtained for in-scene classification and $700$ provided target training examples, where during test time we are given ground-truth bounding boxes and context knowledge about the set of objects present in the image. The goal of the algorithm is then to assign the weak labels to the given bounding-boxes. 
    With this type of scene knowledge and by only considering images with more than one category, we obtain an accuracy of $59.21\%$ compared to $57.53\%$ for SVM-Target and $53.14\%$ for SVM-Source.
    In contrast to \cite{wang2010comparative}, we are not given the exact number of objects for each category in the image, making our problem setting more difficult and realistic. 

  \subsection{Transferring new category models}
  \label{sec:newcategories}

    \begin{figure}
      \centering
      \includegraphics[angle=270,width=0.7\linewidth]{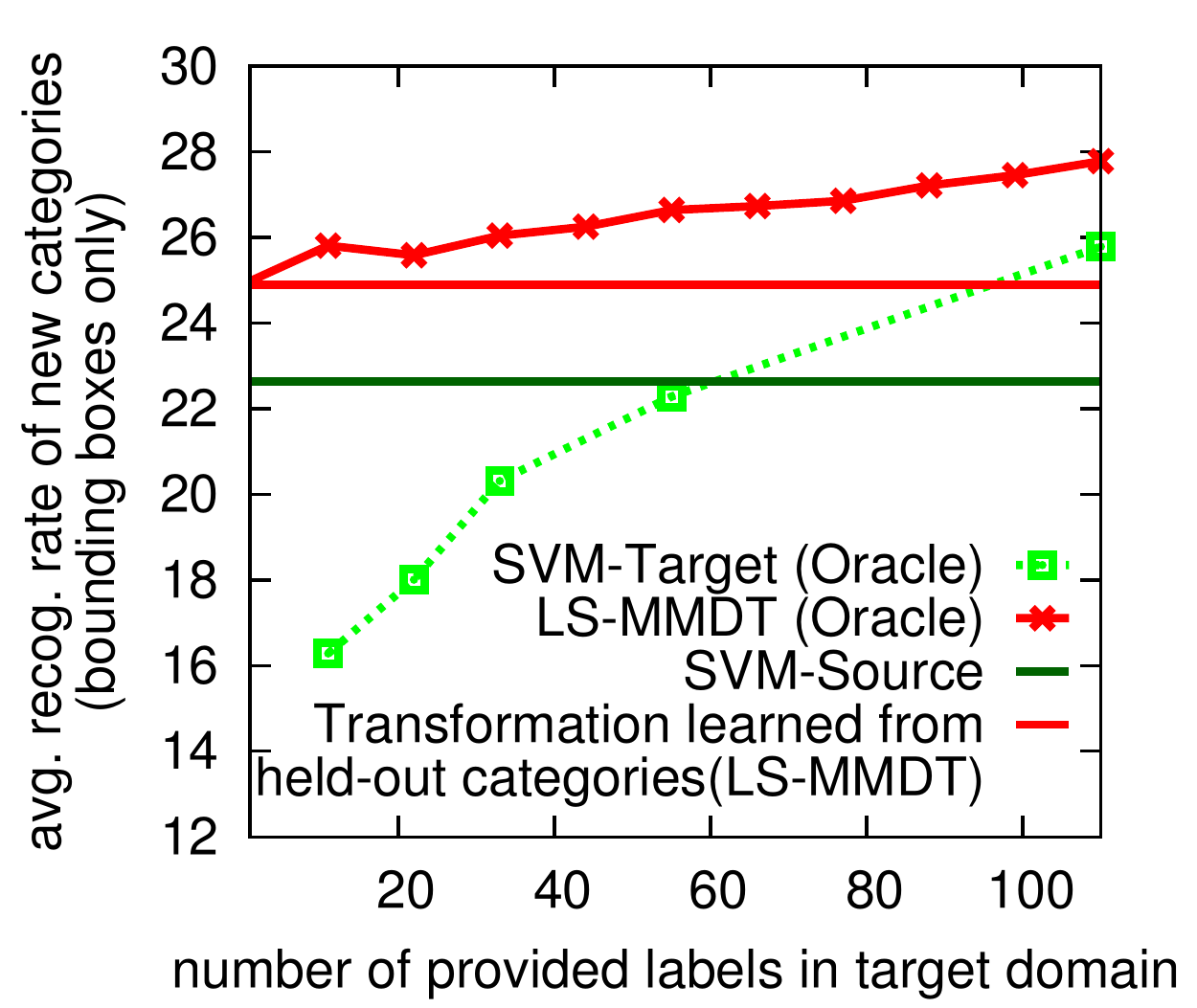}
      \includegraphics[angle=270,width=0.25\linewidth]{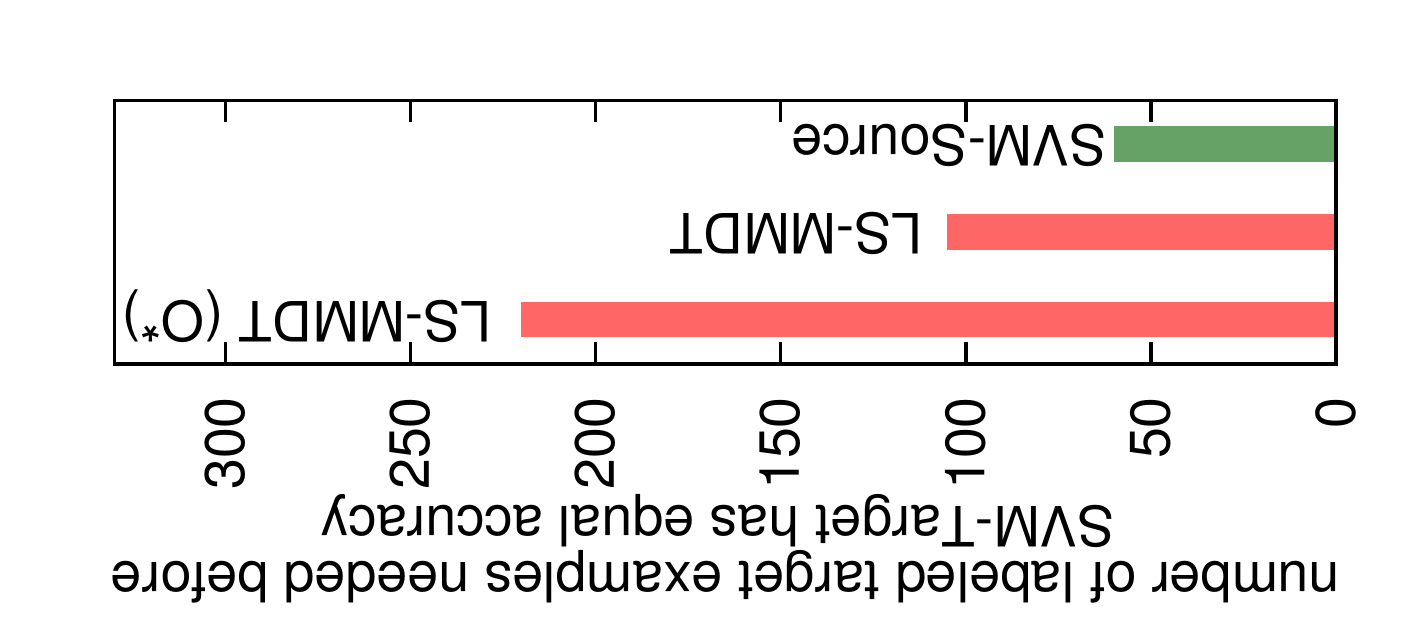}
      \caption{New category scenario: our approach is used to learn a transformation from held-out categories and to transfer new category models directly from the source domain without target examples. The performance is compared to an oracle SVM-Target and MMDT that use target examples from the held-out categories.} 
      \label{fig:newcat}
      \vspace{-.4cm}
    \end{figure}

    A key benefit of our method is the possibility of transferring category models to the target domain even when no target domain examples are available at all. 
    In the following experiment, we selected $11$ categories\footnote{laptop,phone,toaster,keyboard,fan,printer,teapot,chair,basket,clock,bottle} from our ImageNet/SUN2012 dataset and only provided training examples in the source domain for them. The transformation is learned from all other categories with both labeled examples in the target and the source domain. 

    As we can see in \figurename~\ref{fig:newcat}, this transfer method (``Transf. learned from other categories'') even outperforms learning in the target domain (SVM-Target Oracle) with up to 100 labeled training examples. Especially with large-scale datasets, like ImageNet, this ability of our fast transform-based adaptation method provides a huge advantage and allows using all visual categories provided in the source as well as in the target domain. Furthermore, the experiment shows that we indeed learn a category-invariant transformation that can compensate for the observed dataset bias~\cite{ref:Efros-dataset-bias-cvpr2011}.

  \subsection{Adapting from different feature types}
  \label{sec:diffdim}

    Transform-based domain adaptation can be also applied when source and target domain have different feature dimensionality. To show the applicability of our method in this setting we use the same setup as in the previous experiment, but we computed $1500$-dimensional BoW features for objects in the SUN2012 dataset and learned a transformation from the $1000$ dimensional features in the ImageNet dataset. Adaptation with our approach achieves a recognition rate of $18.2\%$ compared to $16.9\%$ of SVM-Target using one target training example per category.
    This can be seen as one of the most difficult adaptation scenarios, where we estimate the domain transformation from different categories and between completely different feature spaces.

\vspace{-0.235cm}

\section{Conclusions}

  In this paper, we showed how to extend transform-based domain adaptation towards large-scale scenarios. Our method allows for efficient estimation of a category-invariant domain transformation in the cases of large feature dimensionality and a large number of training examples. This is done by exploiting an implicit low-rank structure of the transformation and by making explicit use of a close connection to standard max-margin problems and efficient optimization techniques for them. Our method is easy to implement and apply, and achieves significant performance gains when adapting visual recognition models learned from biased internet sources to real-world scene understanding datasets.

  An important take-home message of this paper is that collecting more and more annotated visual data does not necessarily help for solving scene understanding in general. However, domain adaptation can help to bridge the gap by learning category-invariant transformations without significant additional computational overhead.

{\small
\bibliographystyle{ieee}
\bibliography{largescale-da}
}

\end{document}